\documentclass[10pt,twocolumn,letterpaper]{article}

\usepackage[pagenumbers]{cvpr} 

\usepackage{graphicx}
\usepackage{amsmath}
\usepackage{amssymb}
\usepackage{booktabs}
\usepackage{enumitem}
\usepackage[misc]{ifsym}

\usepackage[pagebackref,breaklinks,colorlinks]{hyperref}

\usepackage[capitalize]{cleveref}
\crefname{section}{Sec.}{Secs.}
\Crefname{section}{Section}{Sections}
\Crefname{table}{Table}{Tables}
\crefname{table}{Tab.}{Tabs.}

\def\cvprPaperID{2273} 
\def\confName{CVPR}
\def\confYear{2023}

\begin{document}

\title{Rerender A Video: Zero-Shot Text-Guided Video-to-Video Translation\vspace{-3mm}}

\author{Shuai Yang \hspace{12pt} Yifan Zhou  \hspace{12pt}  Ziwei Liu  \hspace{12pt} Chen Change Loy$^{~\textrm{\Letter}}$\\
S-Lab, Nanyang Technological University\\
{\tt\small \{shuai.yang, yifan.zhou,  ziwei.liu, ccloy\}@ntu.edu.sg}\vspace{-1mm}
}

\twocolumn[{%
\renewcommand\twocolumn[1][]{#1}%
\maketitle
\vspace{-2.3em}
\begin{center}
\centering
\includegraphics[width=\linewidth]{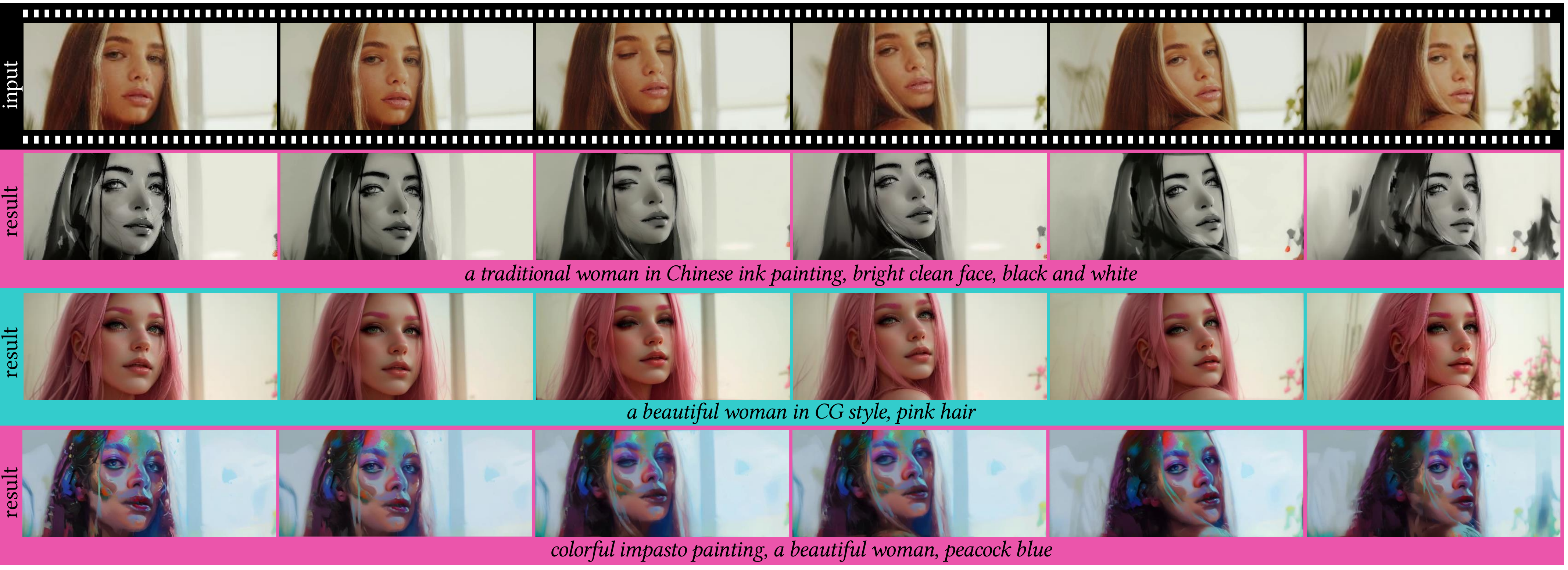}
\captionof{figure}{We present a novel video-to-video translation framework that can render a source input video into a temporal-coherent video with the style specified by a target textual description.}
\label{fig:teaser}
\end{center}%
}]

\begin{abstract}
   Large text-to-image diffusion models have exhibited impressive proficiency in generating high-quality images. However, when applying these models to video domain, ensuring temporal consistency across video frames remains a formidable challenge. This paper proposes a novel zero-shot text-guided video-to-video translation framework to adapt image models to videos. The framework includes two parts: key frame translation and full video translation. The first part uses an adapted diffusion model to generate key frames, with hierarchical cross-frame constraints applied to enforce coherence in shapes, textures and colors. The second part propagates the key frames to other frames with temporal-aware patch matching and frame blending. Our framework achieves global style and local texture temporal consistency at a low cost (without re-training or optimization). The adaptation is compatible with existing image diffusion techniques, allowing our framework to take advantage of them, such as customizing a specific subject with LoRA, and introducing extra spatial guidance with ControlNet. Extensive experimental results demonstrate the effectiveness of our proposed framework over existing methods in rendering high-quality and temporally-coherent videos. Code is available at our project page:
\url{https://www.mmlab-ntu.com/project/rerender/}
\end{abstract}

\section{Introduction}

Recent text-to-image diffusion models such as DALLE-2~\cite{ramesh2022hierarchical}, Imagen~\cite{saharia2022photorealistic}, Stable Diffusion~\cite{rombach2022high} demonstrate exceptional ability in generating diverse and high-quality images guided by natural language.
Based on it, a multitude of image editing methods have emerged, including model fine-tuning for customized object generation~\cite{ruiz2022dreambooth}, image-to-image translation~\cite{meng2021sdedit}, image inpainting~\cite{avrahami2022blended}, and object editing~\cite{hertz2022prompt}.
These applications allow users to synthesize and edit images effortlessly, using natural language within a unified diffusion framework, greatly improving creation efficiency.
As video content surges in popularity on social media platforms, the demand for more streamlined video creation tools has concurrently risen.
Yet, a critical challenge remains: the direct application of existing image diffusion models to videos leads to severe flickering issues.

Researchers have recently turned to text-guided video diffusion models and proposed three solutions.
The first solution involves training a video model on large-scale video data~\cite{ho2022imagen}, which requires significant computing resources. Additionally, the re-designed video model is incompatible with existing off-the-shelf image models.
The second solution is to fine-tune image models on a single video~\cite{wu2022tune}, which is less efficient for long videos. Overfitting to a single video may also degrade the performance of the original models.
The third solution involves zero-shot methods~\cite{khachatryan2023text2video} that require no training. During the diffusion sampling process, cross-frame constraints are imposed on the latent features for temporal consistency. The zero-shot strategy requires fewer computing resources and is mostly compatible with existing image models, showing promising potential.
However, current cross-frame constraints are limited to global styles and are unable to preserve low-level consistency, \eg, the overall style may be consistent, but the local structures and textures may still flicker.

\begin{figure}[t]
\centering
\includegraphics[width=\linewidth]{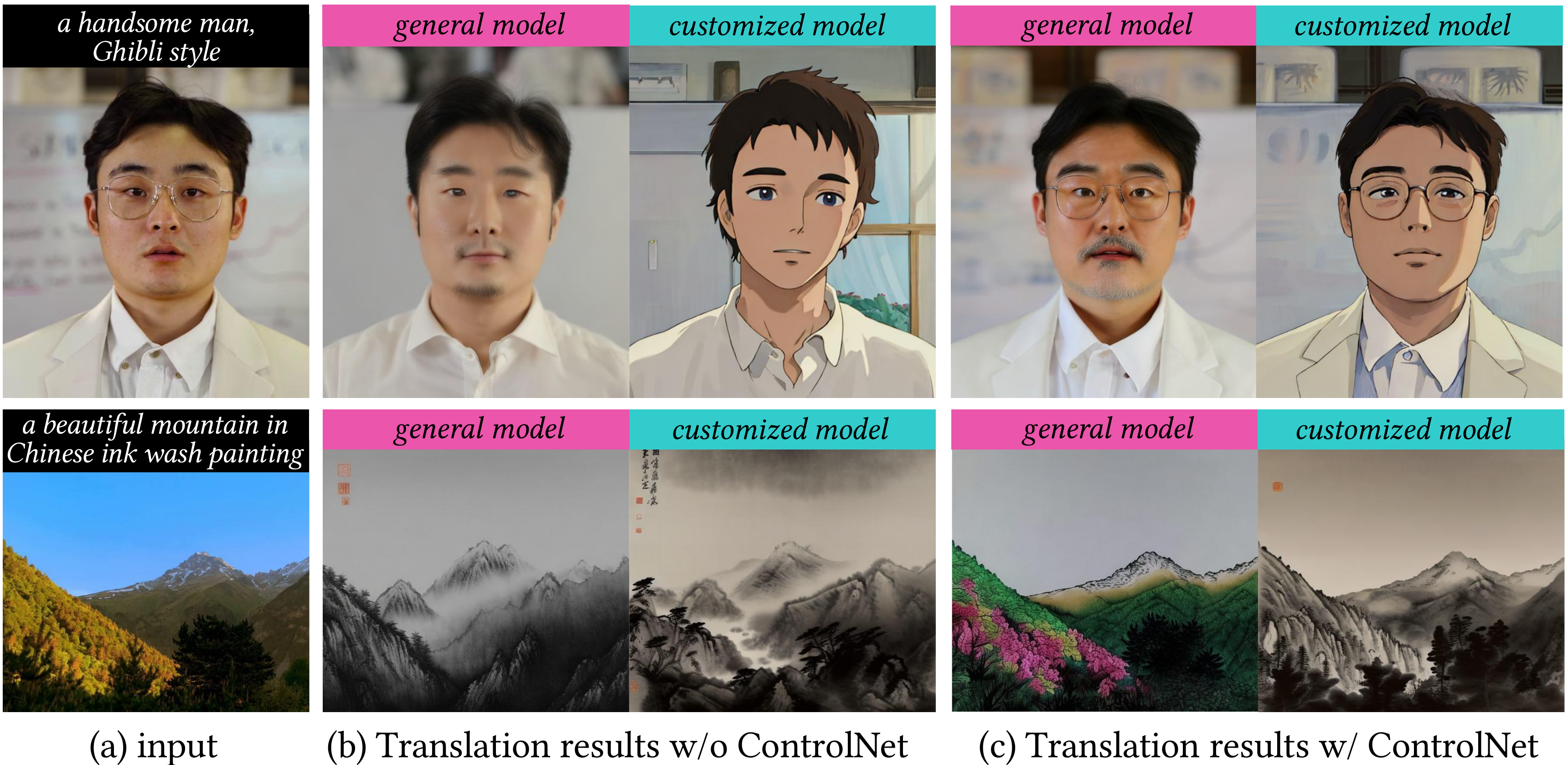}\vspace{-2mm}
\caption{Customized model and ControlNet generate high-quality results with better consistency with both prompt and content. Our method is designed to be compatible with these existing image diffusion techniques, and thus can take advantage of them to strike a good balance between the style (prompt) and the content.}\vspace{-6mm}
\label{fig:zero_shot}
\end{figure}

Achieving successful application of image diffusion models to the video domain is a challenging task. It requires
\textbf{1)} Temporal consistency: cross-frame constraints for low-level consistency;
\textbf{2)} Zero-shot: no training or fine-tuning required;
\textbf{3)} Flexibility: compatible with off-the-shelf image models for customized generation.
As mentioned above, image models can be customized by fine-tuning on specific objects to capture the target style more precisely than general models. Figure~\ref{fig:zero_shot} shows two examples.
To take advantage of it, in this paper, we employ zero-shot strategy for model compatibility and aim to further solve the key issue of this strategy in maintaining low-level temporal consistency.

To achieve this goal, we propose novel hierarchical cross-frame constraints for pre-trained image models to produce coherent video frames.
Our key idea is to use optical flow to apply dense cross-frame constraints,
with the previous rendered frame serving as a low-level reference for the current frame and the first rendered frame acting as an anchor to regulate the rendering process to prevent deviations from the initial appearance.
Hierarchical cross-frame constraints are realized at different stages of diffusion sampling.
In addition to global style consistency, our method enforces consistency in shapes, textures and colors at early, middle and late stages, respectively.
This innovative and lightweight modification achieves both global and local temporal consistency. Figure~\ref{fig:teaser} presents our coherent video translation results over off-the-shelf image models customized for six unique styles.

Based on the insight, this paper introduces a novel zero-shot framework for text-guided video-to-video translation, consisting of two parts: key frame translation and full video translation.
In the first part, we adapt pre-trained image diffusion models with hierarchical cross-frame constraints for generating key frames.
In the second part, we propagate the rendered key frames to other frames using temporal-aware patch matching and frame blending.
The diffusion-based generation is excellent at content creation, but its multi-step sampling process is inefficient.
The patch-based propagation, on the other hand, can efficiently infer pixel-level coherent frames but is not capable of creating new content. By combining these two parts, our framework strikes a balance between quality and efficiency.
To summarize, our main contributions are as follows:
\begin{itemize}[itemsep=1.5pt,topsep=2pt,parsep=1.5pt]
  \item A novel zero-shot framework for text-guided video-to-video translation, which achieves both global and local temporal consistency, requires no training, and is compatible with pre-trained image diffusion models.
  \item Hierarchical cross-frame consistency constraints to enforce temporal consistency in shapes, textures and colors, which adapt image diffusion models to videos.
  \item Hybrid diffusion-based generation and~patch-based propagation to strike a balance between quality and efficiency.
\end{itemize}

\section{Related Work}

\subsection{Text Driven Image Generation}

Generating images with descriptive sentences is intuitive and flexible.
Early attempts explore GAN~\cite{zhang2017stackgan,xu2018attngan,zhu2019dm,zhang2021cross} to synthesize realistic images.
With the powerful expressivity of Transformer~\cite{vaswani2017attention}, autoregressive models~\cite{ramesh2021zero,gafni2022make,ding2021cogview} are proposed to model image pixels as a sequence with autoregressive dependency between each pixel. DALL-E~\cite{ramesh2021zero} and CogView~\cite{ding2021cogview} train an autoregressive transformer on image and text tokens. Make-A-Scene~\cite{gafni2022make} further considers segmentation masks as condition.

Recent studies focus on diffusion models~\cite{ho2020denoising} for text-to-image generation, where images are synthesized via a gradual denoising process.
DALLE-2~\cite{ramesh2022hierarchical} and Imagen~\cite{saharia2022photorealistic} introduce pretrained large language models~\cite{raffel2020exploring,radford2021learning} as text encoder to better align the image with text, and cascade diffusion models for high resolution image generation.
GLIDE~\cite{nichol2022glide} introduces classifier-free guidance to improve text conditioning.
Instead of applying denoising in the image space, Latent Diffusion Models~\cite{rombach2022high} uses the low-resolution latent space of VQ-GAN~\cite{esser2021taming} to improve the efficiency. We refer to~\cite{croitoru2023diffusion} for a thorough survey.

In addition to diffusion models for general images, customized models are studied.
Textual Inversion~\cite{gal2022image} and DreamBooth~\cite{ruiz2022dreambooth} learn special tokens to capture novel concepts and generate related images given a small number of example images.
LoRA~\cite{hulora} accelerates the fine-tuning large models by learning low-rank weight matrices added to existing weights.
ControlNet~\cite{zhang2023adding} fine-tunes a new control path to provide pixel-level conditions such as edge maps and pose, enabling fine-grained image generation. Our method does not alter the pre-trained model, thus is orthogonal to these existing techniques. This empowers our method to leverage DreamBooth and LoRA for better customized video translation and to use ControlNet for temporal-consistent structure guidance as in Fig.~\ref{fig:zero_shot}.

\subsection{Video Editing with Diffusion Models}

For text-to-video generation, Video Diffusion Model~\cite{hovideo} proposes to extend the 2D U-Net in image model to a factorized space-time UNet. Imagen Video~\cite{ho2022imagen} scales up the Video Diffusion Model with a cascade of spatial and temporal video super-resolution models, which is further extended to video editing by Dreamix~\cite{molad2023dreamix}.
Make-A-Video~\cite{singer2022make} leverages video data in an unsupervised manner to learn the movement to drive the image model.
Although promising, the above methods need large-scale video data for training.

Tune-A-Video~\cite{wu2022tune} instead inflates an image diffusion model into a video model with cross-frame attention, and fine-tunes it on a single video to generate videos with related motion. Based on it, Edit-A-Video~\cite{shin2023edit}, Video-P2P~\cite{liu2023video} and vid2vid-zero~\cite{wang2023zero} utilize Null-Text Inversion~\cite{mokady2022null} for precise inversion to preserve the unedited region.
However, these models need fine-tuning of the pre-trained model or optimization over the input video, which is less efficient.

Recent developments have seen the introduction of zero-shot methods that, by design, operate without any training phase. Thus, these methods are naturally compatible with pre-trained diffusion variants like InstructPix2Pix~\cite{brooks2022instructpix2pix} or ControlNet to accept more flexible conditions like depth and edges.
Based on the editing masks detected by Prompt2Prompt~\cite{hertz2022prompt} to indicate the channel and spatial region to preserve, FateZero~\cite{qi2023fatezero} blends the attention features before and after editing.
Text2Video-Zero~\cite{khachatryan2023text2video} translates the latent to directly simulate motions and Pix2Video~\cite{ceylan2023pix2video} matches the latent of the current frame to that of the previous frame.
All the above methods largely rely on cross-frame attention and early-step latent fusion to improve temporal consistency. However, as we will show later, these strategies predominantly cater to high-level styles and shapes, and being less effective in maintaining cross-frame consistency at the level of texture and detail.
In contrast to these approaches, our method proposes a novel pixel-aware cross-frame latent fusion, which non-trivially achieves pixel-level temporal consistency.

Another zero-shot solution is to apply frame interpolation to infer the videos based on one or more diffusion-edited frames. The seminal work of image analogy~\cite{Hertzmann2001Image} migrates the style effect from an exemplar pair to other images with patch matching. Fi\v{s}er \textit{et al.}~\cite{fivser2017example} extend image analogy to facial video translation with the guidance of facial features. Later, Jamriv\v{s}ka \textit{et al.}~\cite{jamrivska2019stylizing} propose an improved EbSynth for general video translation based on multiple exemplar frames with a novel temporal blending approach.
Although these patch-based methods can preserve fine details, their temporal consistency largely relies on the coherence across the exemplar frames. Thus, our adapted diffusion model for generating coherent frames is well suited for these methods, as we will show later in Fig.~\ref{fig:ablation-pafusion2}. In this paper, we integrate the zero-shot EbSynth into our framework to achieve better temporal consistency and accelerate inference without any further training.

\section{Preliminary: Diffusion Models}

\paragraph{Stable Diffusion} Stable Diffusion is a latent diffusion model operating in the latent space of an autoencoder $\mathcal{D}(\mathcal{E}(\cdot))$, where $\mathcal{E}$ and  are the encoder and decoder, respectively.
Specifically, for an image $I$ with its latent feature $x_0=\mathcal{E}(I)$, the diffusion forward process iteratively add noises to the latent
\begin{equation}
  q(x_t|x_{t-1})=\mathcal{N}(x_t;\sqrt{\alpha_t}x_{t-1}, (1-\alpha_t)\mathbf{I}),
\end{equation}
where $t=1,...,T$ is the time step, $q(x_t|x_{t-1})$ is the conditional density of $x_t$ given $g_{t-1}$, and $\alpha_t$ is hyperparameters.
Alternatively, we can directly sample $x_t$ at any time step from $x_0$ with,
\begin{equation}\label{eq:forward_sample}
  q(x_t|x_{0})=\mathcal{N}(x_t;\sqrt{\bar{\alpha}_t}x_0, (1-\bar{\alpha}_t)\mathbf{I}),
\end{equation}
where $\bar{\alpha}_t=\prod_{i=1}^t\alpha_i$.

\begin{figure}[t]
\centering
\includegraphics[width=\linewidth]{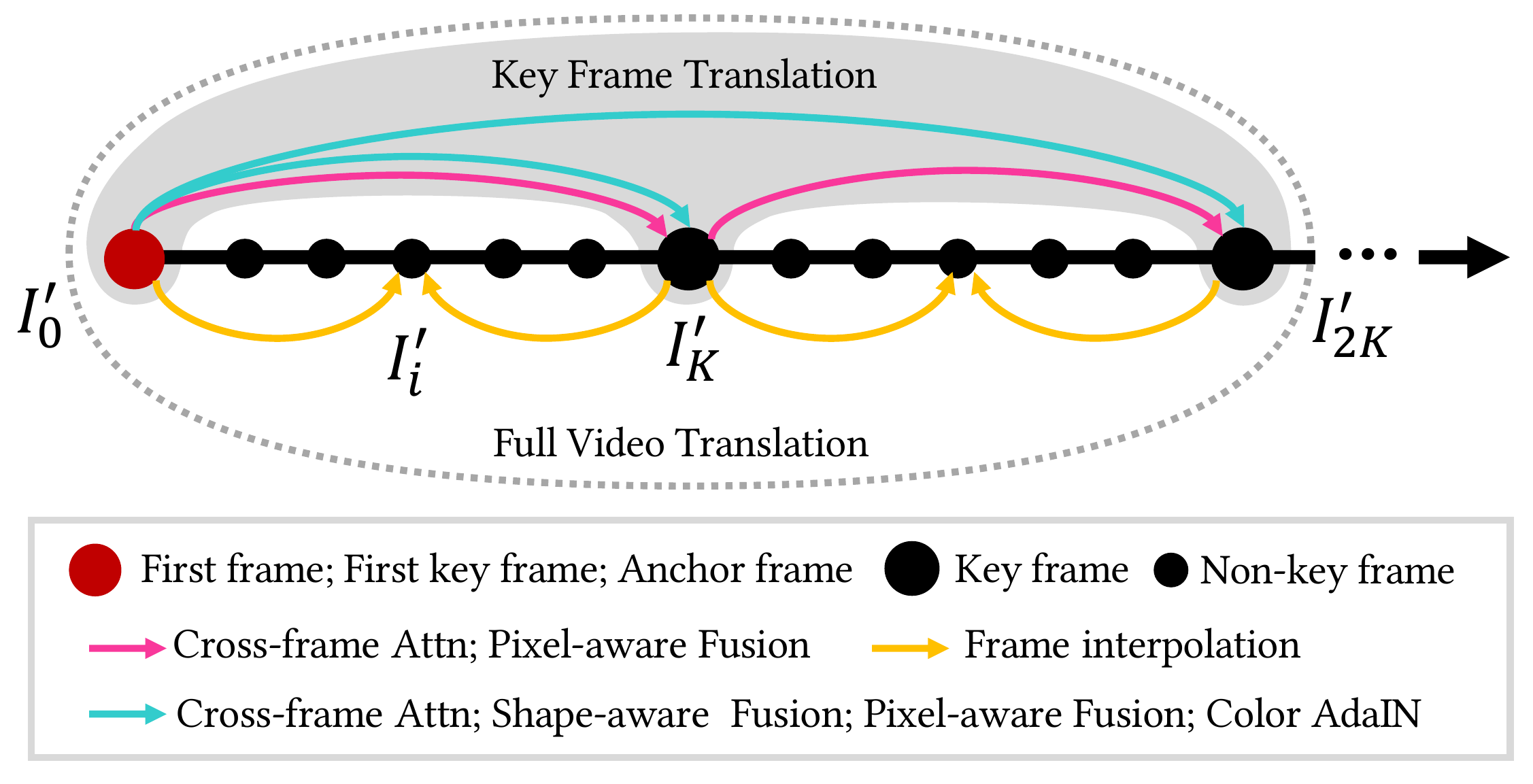}\vspace{-2mm}
\caption{Illustration of the interaction between different frames to impose temporal constraints in our framework.}\vspace{-3mm}
\label{fig:relation}
\end{figure}

\begin{figure*}[htbp]
\centering
\includegraphics[width=\linewidth]{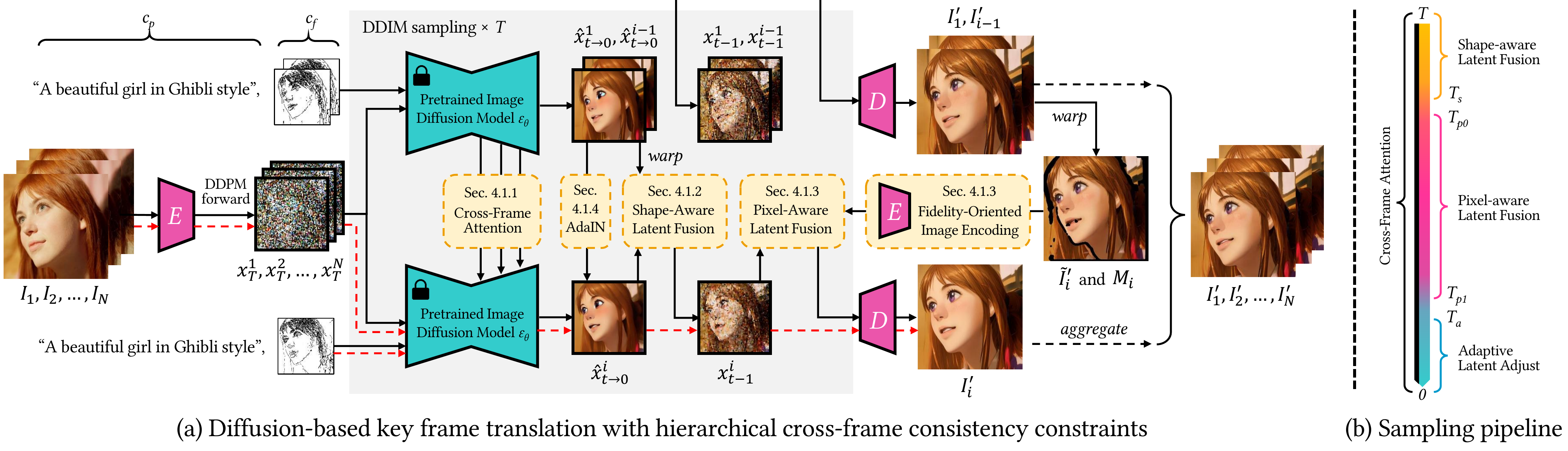}\vspace{-2mm}
\caption{Framework of the proposed zero-shot text-guided video translation. (a) We adapt the pre-trained image diffusion model (Stable Diffusion + ControlNet) with hierarchical cross-frame constraints to render coherent frames. The red dotted lines denote the sampling process of the original image diffusion model. The black lines denote our adapted process for video translation. (b) We apply different constraints at different sampling steps.}\vspace{-3mm}
\label{fig:framework1}
\end{figure*}

Then in the diffusion backward process, a U-Net $\epsilon_\theta$ is trained to predict the noise of the latent to iteratively recover $x_0$ from $x_T$. Given a large $T$, $x_0$ will be completely destroyed in the forward process so that $x_T$ approximates a standard Gaussian distribution. Therefore, $\epsilon_\theta$ correspondingly learns to infer valid $x_0$ from random Gaussian noises.
Once trained, we can sample $x_{t-1}$ based on $x_{t}$ with a deterministic DDIM sampling~\cite{songdenoising}:
\begin{equation}\label{eq:ddim}
  x_{t-1}=\sqrt{\alpha_{t-1}}\underbrace{\hat{x}_{t\rightarrow0}}_{\text{predicted}~x_0}
  +\underbrace{\sqrt{1-\alpha_{t-1}}\epsilon_\theta(x_t, t, c_p)}_{\text{direction pointing to}~x_{t-1}},
\end{equation}
where $\hat{x}_{t\rightarrow0}$ is the predicted $x_0$ at time step $t$,
\begin{equation}
  \hat{x}_{t\rightarrow0}=(x_t-\sqrt{1-\alpha_{t}}\epsilon_\theta(x_t, t, c_p))/\sqrt{\alpha_{t}},
\end{equation}
and $\epsilon_\theta(x_t, t, c_p)$ is the predicted noise of $x_t$  based on the time step $t$ and the text prompt condition $c_p$.

During inference, we can sample a valid $x_0$ from the standard Guassian noise $x_T=z_T, z_T\sim\mathcal{N}(0, \mathbf{I})$ with DDIM sampling, and decode $x_0$ to the final generated image $I'=\mathcal{D}(x_0)$.

\paragraph{ControlNet} Although flexible, natural language has limited spatial control over the output.
To improve spatial controllability, \cite{zhang2023adding} introduce a side path called ControlNet to Stable Diffusion to accept extra conditions like edges, depth and human pose. Let $c_f$ be the extra condition, the noise prediction of U-Net with ControlNet becomes $\epsilon_\theta(x_t, t, c_p, c_f)$. Compared to InstructPix2Pix, ControlNet is orthogonal to customized Stable Diffusion models. To build a general zero-shot V2V framework, we use ControlNet to provide structure guidance from the input video to improve temporal consistency.

\section{Zero-Shot Text-Guided Video Translation}

Given a video with $N$ frames $\{I_i\}_{i=0}^N$, our goal is to render it into a new video $\{I'_i\}_{i=0}^N$ in another artistic expression specified by text prompts and/or off-the-shelf customized Stable Diffusion models. 
Our framework consists of two parts: Key Frame Translation (Sec. \ref{sec:method1}) and Full Video Translation (Sec. \ref{sec:method2}).
In the first part, we introduce four hierarchical cross-frame constraints into pre-trained image diffusion models, guiding the rendering of coherent key frames using anchor and previous key frames, as as illustrated in Fig.~\ref{fig:relation}.
Then in the second part, non-key frames are interpolated based on their neighboring two key frames.
Thus our framework can fully exploit the relationship between different frames to enhance temporal consistency of the outputs.

\begin{table}[]
\begin{center}
\caption{Notation summary.}\vspace{-2mm}
\resizebox{\linewidth}{!}{
\begin{tabular}{c|l}
\toprule
\textbf{Notation} & \textbf{Description} \\
\midrule
$\mathcal{E}$, $\mathcal{D}$ & image encoder and decoder of Stable Diffusion \\
$\mathcal{E}^*$ & the proposed fidelity-oriented image encoder \\
$I_i$ & the $i$-th key frame (Sec.~\ref{sec:method1}); the $i$-th video frame (Sec.~\ref{sec:method2}) \\
$I'_i$ & translation result of $I_i$ with the proposed method \\
$\bar{I}'_{i}$ & translation result of $I_i$ without PA fusion (Fig.~\ref{fig:ablation-pafusion2}(b)) \\
$\tilde{I}'_i$, $M_i$ & result of warping $I'_0$ and $I'_{i-1}$ to $I_i$, and its occlusion mask \\
$I'^{j}_i$ & result of propagating the translated key frame $I'_j$ to $I_i$ \\
$x^i_0$ & latent feature of $I_i$ encoded by $\mathcal{E}$ \\
$x^i_t$ & latent feature of $I_i$ at the diffusion backward denoising step $t$ \\
$\hat{x}^i_{t\rightarrow0}$ & estimated $x^i_0$ of $I_i$ at the diffusion backward denoising step $t$ \\
$\tilde{x}^i_t$ & latent feature of $\tilde{I}'_i$ at the diffusion forward sampling step $t$ \\
$w^i_j$, $M^i_j$ & optical flow and occlusion mask from $I_j$ to $I_i$ \\
\bottomrule
\end{tabular}}\vspace{-3mm}
\label{tb:notation}
\end{center}
\end{table}

\subsection{Key Frame Translation}\label{sec:method1}

Figure~\ref{fig:framework1} illustrates the $T$-step sampling pipeline for the key frame translation.
Following SDEdit~\cite{meng2021sdedit}, the pipeline begins with $x_T=\sqrt{\bar{\alpha}_T}x_0+(1-\bar{\alpha}_T)z_T, z_T\sim\mathcal{N}(0, \mathbf{I})$, the noisy latent code of the input video frame rather than the pure Gaussian noise.
It enables users to determine how much detail of the input frame is preserved in the output by adjusting $T$, \ie, smaller $T$ retain more detail. Then, during sampling each frame, we use the first frame as anchor frame and its previous frame to constrain global style consistency and local temporal consistency.

Specifically, cross-frame attention~\cite{wu2022tune} is applied to all sampling steps for global style consistency (Sec.~\ref{sec:method1-1}).
In addition, in early steps, we fuse the latent feature with the aligned latent feature of previous frame to achieve rough shape alignments (Sec.~\ref{sec:method1-2}). Then in mid steps, we use the latent feature with the encoded warped anchor and previous outputs to realize fine texture alignments (Sec.~\ref{sec:method1-3}). Finally, in late steps, we adjust the latent feature distribution for color consistency (Sec.~\ref{sec:method1-4}).
For simplicity, we will use $\{I_i\}_{i=0}^N$ to refer to the key frames in this section. We summarize important notations in Table~\ref{tb:notation}.

\subsubsection{Style-aware cross-frame attention}\label{sec:method1-1}

Similar to other zero-shot video editing methods~\cite{khachatryan2023text2video,ceylan2023pix2video}, we replace self-attention layers in the U-Net with cross-frame attention layers to regularize the global style of $I'_i$ to match that of $I'_1$ and $I'_{i-1}$.
In Stable Diffusion, each self-attention layer receives the latent feature $v_i$ (for simplicity we omit the time step $t$) of $I_i$, and linearly projects $v_i$ into query, key and value $Q$, $K$, $V$ to produce the output by $\textit{Self\_Attn}(Q,K,V)=\textit{Softmax}(\frac{QK^T}{\sqrt{d}})\cdot V$ with
\begin{equation}
  Q=W^Qv_i, K=W^Kv_i, V=W^Vv_i,
\end{equation}
where $W^Q$, $W^K$, $W^V$ are pre-trained matrices for feature projection.
Cross-frame attention, by comparison, uses the key $K'$ and value $V'$ from other frames (we use the first and previous frames), \ie,
$\textit{CrossFrame\_Attn}(Q,K',V')=\textit{Softmax}(\frac{QK'^T}{\sqrt{d}})\cdot V'$ with
\begin{equation}
  Q=W^Qv_i, K'=W^K[v_1;v_{i-1}], V'=W^V[v_1;v_{i-1}].
\end{equation}

Intuitively, self-attention can be thought as patch matching and voting within a single frame, while cross-frame attention seeks similar patches and fuses the corresponding patches from other frames, meaning the style of $I'_i$ will inherit that of $I'_1$ and $I'_{i-1}$.

\subsubsection{Shape-aware cross-frame latent fusion}\label{sec:method1-2}

Cross-frame attention is limited to global style. To constrain the cross-frame local shape and texture consistency, we use optical flow to warp and fuse the latent features. Let $w^i_j$ and $M^i_j$ denote the optical flow and occlusion mask from $I_j$ to $I_i$, respectively. Let $x_t^i$ be the latent feature for $I'_i$ at time step $t$. We update the predicted $\hat{x}_{t\rightarrow0}$ in Eq.~(\ref{eq:ddim}) by
\begin{equation}
  \hat{x}^i_{t\rightarrow0}\leftarrow M_{j}^i\cdot\hat{x}^i_{t\rightarrow0} + (1-M_{j}^i)\cdot w_{j}^i(\hat{x}^{j}_{t\rightarrow0}).
\end{equation}
$w$ and $M$ are downsampled to match the resolution of $x$ (we omit the downsampling operation for simplicity in this paper).
For the reference frame $I_j$, we experimentally find that the anchor frame ($j=0$) provides better guidance than the previous frame ($j=i-1$). We observe that interpolating elements in the latent space can lead to blurring and shape distortion in the late steps. Therefore, we limit the fusion to only early steps for rough shape guidance.

\begin{figure}[t]
\centering
\includegraphics[width=\linewidth]{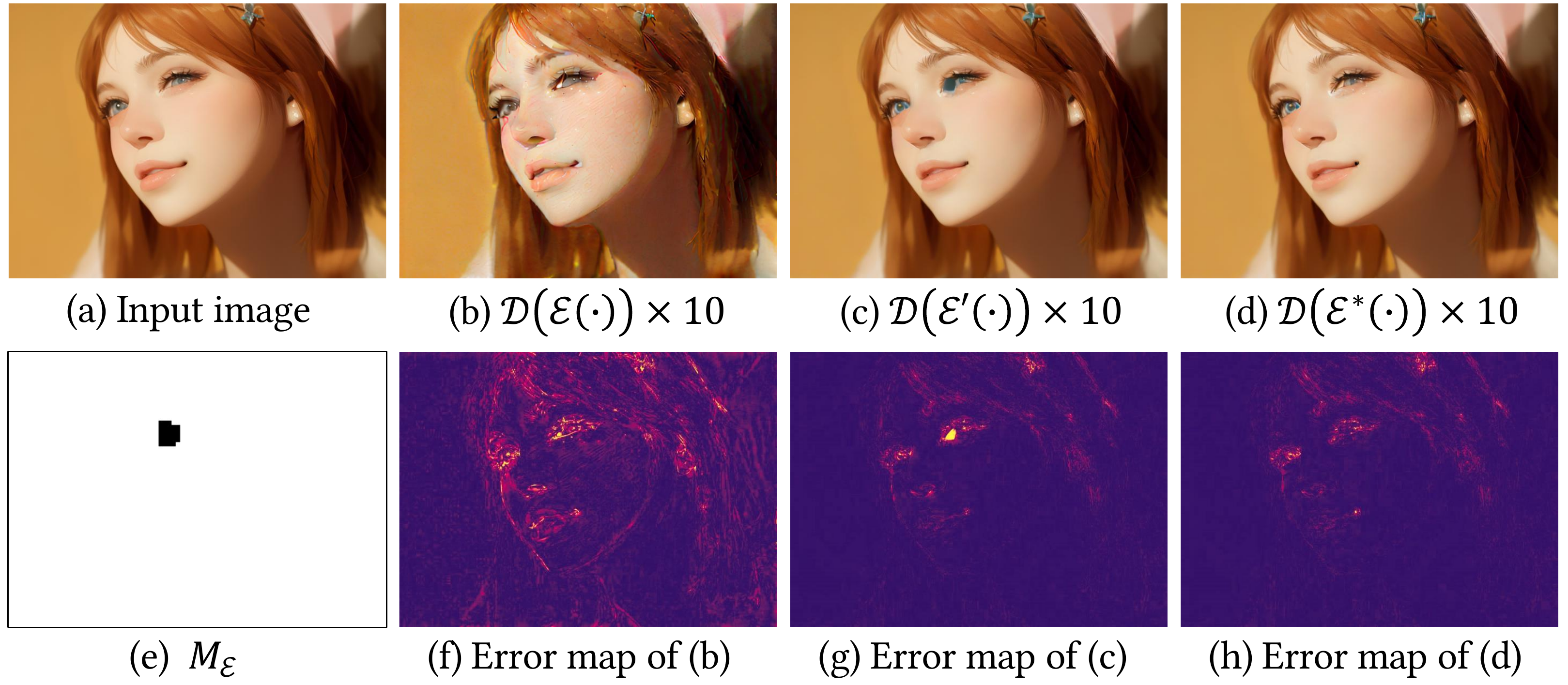}\vspace{-2mm}
\caption{Fidelity-oriented image encoding.}\vspace{-4mm}
\label{fig:encoding}
\end{figure}

\begin{figure}[t]
\centering
\includegraphics[width=\linewidth]{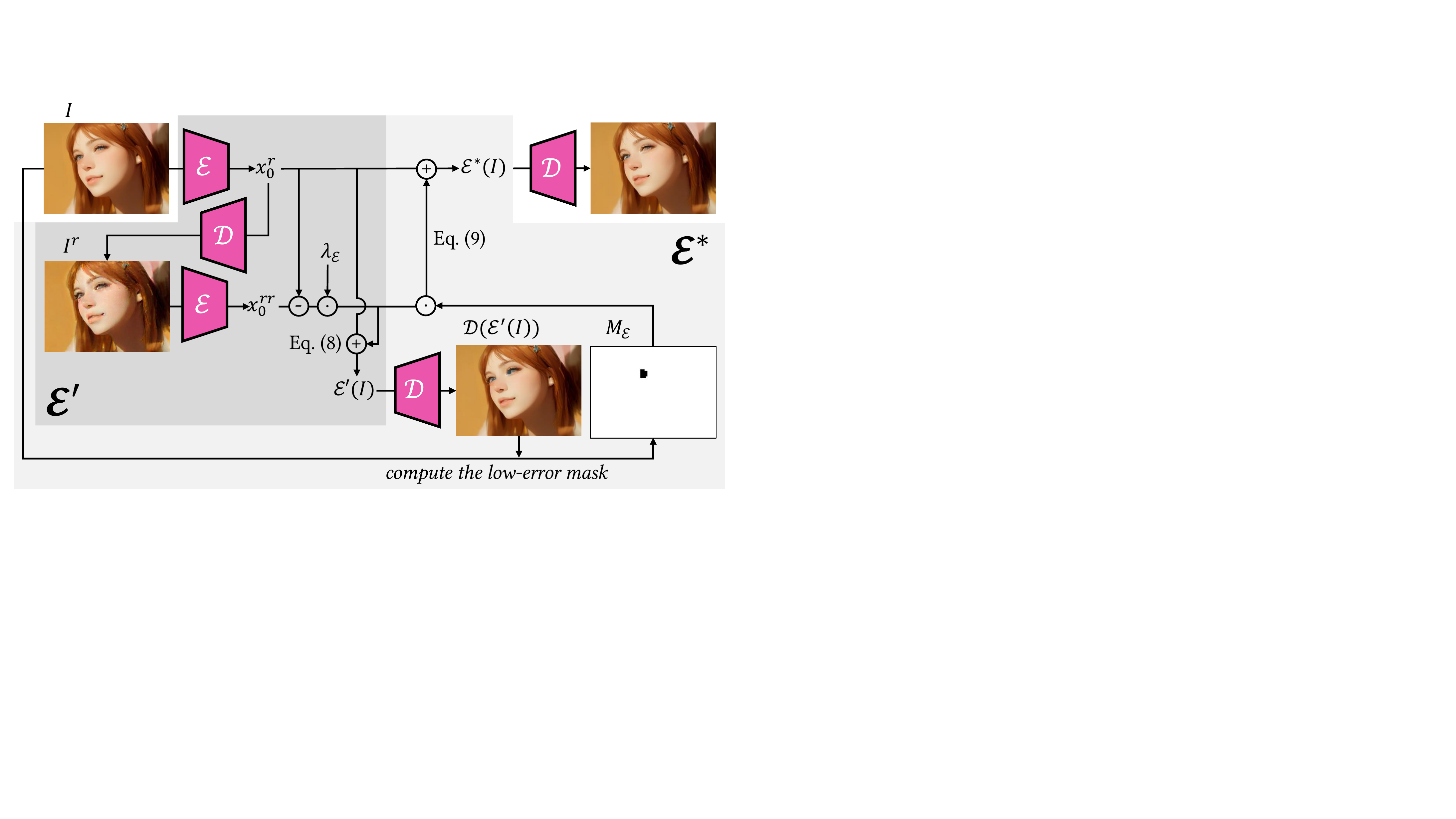}\vspace{-2mm}
\caption{Pipeline of the fidelity-oriented image encoding.}\vspace{-4mm}
\label{fig:pipeline_encoding}
\end{figure}

\subsubsection{Pixel-aware cross-frame latent fusion}\label{sec:method1-3}

To constrain the low-level texture features in mid steps, instead warping the latent feature, we can alternatively warp previous frames and encode them back to the latent space for fusion in an inpainting manner. However, the lossy autoencoder introduces distortions and color bias that easily accumulate along the frame sequence. Figure~\ref{fig:encoding}(b) shows an example of the distorted result after encoding and decoding 10 times. \cite{avrahami2022blended} solved this problem by fine-tuning the decoder's weights to fit each image, which is impractical for long videos. To efficiently solve this problem, we propose a novel fidelity-oriented zero-shot image encoding method.

\paragraph{Fidelity-oriented image encoding} Our key insight is the observation that the amount of information lost each time in the iterative auto-encoding process is consistent. Therefore, we can predict the information loss for compensation. Specifically, for arbitrary image $I$,
we encode and decode it twice, obtaining $x^r_0=\mathcal{E}(I), I_r=\mathcal{D}(x^r_0)$ and $x^{rr}_0=\mathcal{E}(I_r), I_{rr}=\mathcal{D}(x^{rr}_0)$. We assume the loss from the target lossless $x_0$ to $x^r_0$ is linear to that from $x^r_0$ to $x^{rr}_0$.
Then we define the encoding $\mathcal{E}'$ with compensation as
 \begin{equation}
  \mathcal{E}'(I):=x^r_0 +\lambda_\mathcal{E} (x^r_0 - x^{rr}_0),
\end{equation}
where we find the linear coefficient $\lambda_\mathcal{E}=1$ works well. We further add a mask $M_\mathcal{E}$ to prevent the possible artifacts introduced by compensation (\eg, blue artifact near the eyes in Fig.~\ref{fig:encoding}(c)). $M_\mathcal{E}$  indicates where the error between $I$ and $\mathcal{D}(\mathcal{E}'(I))$
is under a pre-defined threshold. Then, our novel fidelity-oriented image encoding $\mathcal{E}^*$ takes the form of
 \begin{equation}
  \mathcal{E}^*(I):=x^r_0 +M_\mathcal{E}\cdot\lambda_\mathcal{E} (x^r_0 - x^{rr}_0).
\end{equation}
The encoding pipeline is summarized in Fig.~\ref{fig:pipeline_encoding}. As shown in Fig.~\ref{fig:encoding}(d), our method preserves image information well even after encoding and decoding 10 times.

\begin{figure}[t]
\centering
\includegraphics[width=\linewidth]{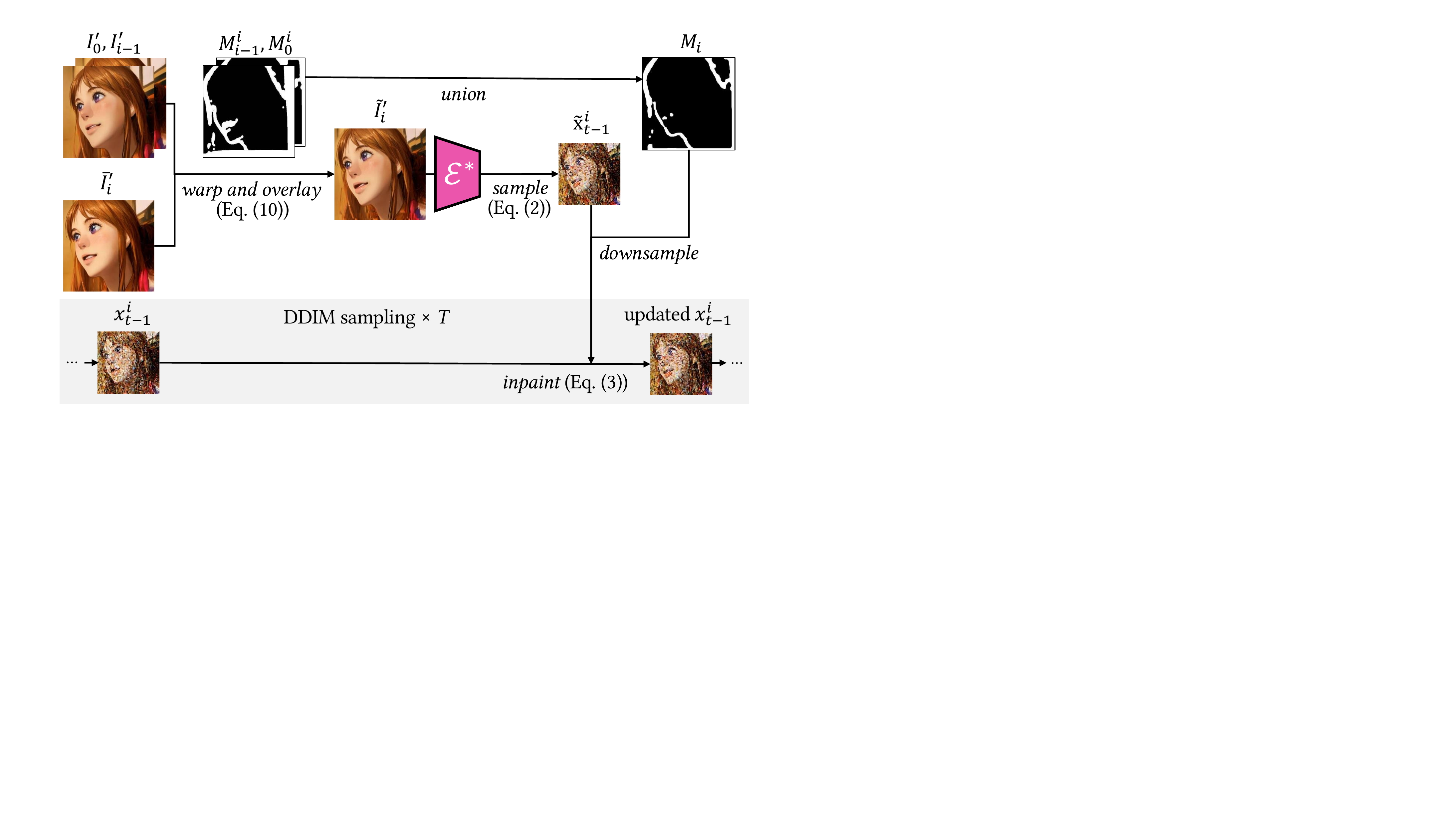}\vspace{-2mm}
\caption{Pipeline of the pixel-aware latent fusion.}\vspace{-4mm}
\label{fig:pipeline_pafusion}
\end{figure}

\paragraph{Structure-guided inpainting} As illustrated in Fig.~\ref{fig:pipeline_pafusion}, for pixel-level coherence, we warp the anchor frame $I'_0$ and the previous frame $I'_{i-1}$ to the $i$-th frame and overlay them on a rough rendered frame $\bar{I}'_{i}$ obtained without the pixel-aware cross-frame latent fusion as
 \begin{equation}\label{eq:warp}
  M_0^i\cdot\big(M_{i-1}^i\cdot\bar{I}'_{i} + (1-M_{i-1}^i)\cdot w^i_{i-1}(I'_{i-1})\big)+\big(1-M_0^i\big)\cdot w_0^i(I'_0)
\end{equation}
The resulting fused frame $\tilde{I}'_i$ provides pixel reference for the sampling of $I'_i$, \ie, we would like $I'_i$ to match $\tilde{I}'_i$ outside the mask area $M_i=M_{0}^i\cap M_{i-1}^i$ and to match the structure guidance from ControlNet inside $M_i$.
We formulate it as a structure-guided inpainting task and follow \cite{avrahami2022blended} to update $x_{t-1}^i$ in Eq.~(\ref{eq:ddim}) as
\begin{equation}
  x^i_{t-1}\leftarrow M_i\cdot x^i_{t-1}+(1-M_i)\cdot \tilde{x}^i_{t-1},
\end{equation}
where $\tilde{x}^i_{t-1}$ is the sampled $x_{t-1}$ from $x_0=\mathcal{E}^*(\tilde{I}'_i)$ based on Eq.~(\ref{eq:forward_sample}).

\subsubsection{Color-aware adaptive latent adjustment}\label{sec:method1-4}

Finally, we apply AdaIN~\cite{huang2017adain} to $\hat{x}^i_{t\rightarrow0}$ to match its channel-wise mean and variance to $\hat{x}^1_{t\rightarrow0}$ in the late steps. It can further keep the color style coherent throughout the whole key frames.

\subsection{Full Video Translation}\label{sec:method2}

For frames with similar content, existing frame interpolation methods like Ebsynth ~\cite{jamrivska2019stylizing} can generate plausible results by propagating the rendered frames to their neighbors efficiently. However, compared to diffusion models, frame interpolation cannot create new content. To balance between quality and efficiency, we propose a hybrid framework to render key frames and other frames with the adapted diffusion model and Ebsynth, respectively.

Specifically, we sample the key frames uniformly for every $K$ frame, \ie, $I_0, I_K, I_{2K}, ...$ and render them to $I'_0, I'_K, I'_{2K}, ...$ by our adapted diffusion model. We then render the remaining non-key frames. Taking $I_i$ ($0<i<K$) for example, we adopt Ebsynth to interpolate $I'_i$ with its neighboring stylized key frames $I'_{0}$ and $I'_{K}$.
Ebsynth has two steps of frame propagation and frame blending. In the following, we will briefly introduce the main idea of these two steps and discuss how we adapt Ebsynth to our framework. For implementation details, please refer to \cite{jamrivska2019stylizing}.

\subsubsection{Single key frame propagation}\label{sec:method2-1}

Frame propagation aims to warp the stylized key frame to its neighboring non-key frames based on their dense correspondences.
We directly follow Ebsynth to adopt a guided path-matching algorithm with color, positional, edge, and temporal guidance for dense correspondence prediction and frame warping. Our framework propagates each key frame to its preceding $K-1$ and succeeding $K-1$ frames. We denote the result of propagating $I'_j$ to $I_i$ as $I'^{j}_i$. For $I_i$ ($0<i<K$), we will obtain two results $I'^0_i$ and $I'^K_i$ from its nearby key frames $I'_{0}$ and $I'_{K}$.

\subsubsection{Temporal-aware blending}\label{sec:method2-2}

Frame blending aims to blend $I'^0_i$ and $I'^K_i$ to a final result $I'_i$. Ebsynth proposes a three-step blending scheme: 1) Combining colors and gradients of $I'^0_i$ and $I'^K_i$ by selecting the ones with lower errors during patch matching (Sec.~\ref{sec:method2-1}) for each location; 2) Using the combined color image as a histogram reference for contrast-preserving blending~\cite{heitz2018high} over $I'^0_i$ and $I'^K_i$ to generate an initial blended image; 3) Employing the combined gradient as a gradient reference for screened Poisson blending~\cite{darabi2012image} over the initial blended image to obtain the final result.
Differently, our framework only adopts the first two blending steps and uses the initial blended image as $I'_i$. We do not apply Poisson blending, which we find sometimes causes artifacts in non-flat regions and is relatively time-consuming.

\begin{figure*}[t]
\centering
\includegraphics[width=\linewidth]{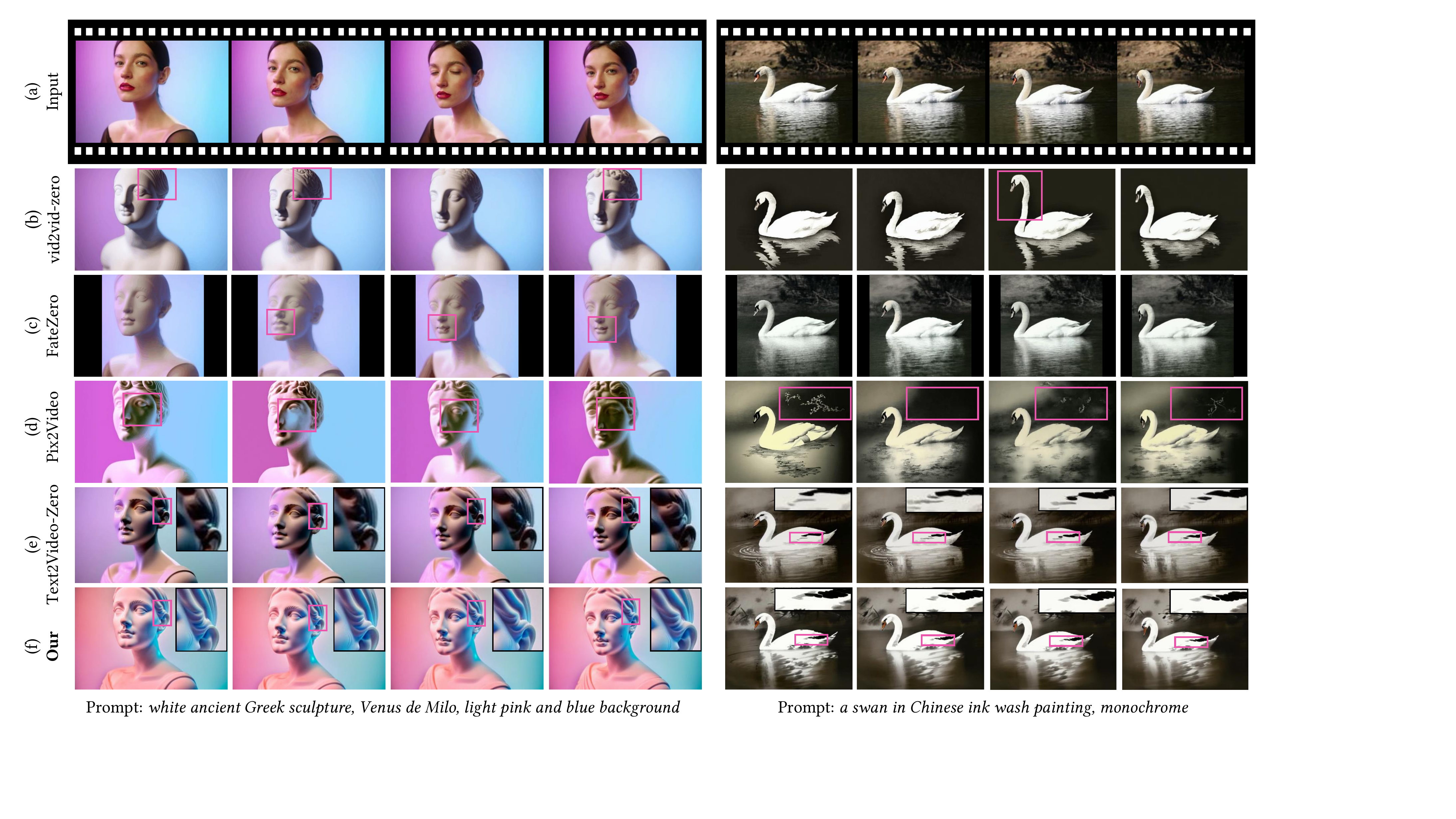}\vspace{-2mm}
\caption{Visual comparison with zero-shot video translation methods. The magenta box indicates the inconsistent region. For Text2Video-Zero and our method, we further enlarge the region to better visualize the pixel-level consistency.}\vspace{-4mm}
\label{fig:compare}
\end{figure*}

\section{Experimental Results}

\subsection{Implementation Details}

The experiment is conducted on one NVIDIA Tesla V100 GPU.
We employ the fine-tuned and LoRA models  based on Stable Diffusion 1.5 from \url{https://civitai.com/}. We use
Stable Diffusion originally uses $T_{max}=1000$ steps. For the sampling pipeline in Fig.~\ref{fig:framework1}(b), by default, we set $T_s=0.1T_{max}$, $T_{p0}=0.5T_{max}$, $T_{p1}=0.8T_{max}$ and $T_a=0.8T_{max}$ and use 20 steps of DDIM sampling. We tune $T$ for each video. ControlNet~\cite{zhang2023adding} is used to provide structure guidance in terms of edges, with the control weight tuned for each video.
We use GMFlow~\cite{xu2022gmflow} for optical flow estimation and compute the occlusion masks by forward-backward consistency check. For full video translation, by default, we sample key frames for every $K=10$ frames. The testing videos are from \url{https://www.pexels.com/} and \url{https://pixabay.com/}, with their short side resized to 512.

In terms of running time for 512$\times$512 videos, key frame and non-key frame translations take about 14.23s and 1.49s per frame, respectively.
Overall, a full video translation takes about $(14.23+1.49(K-1))/K=1.49+12.74/K$s per frame.

We will release our code upon publication of the paper.

\subsection{Comparison with State-of-the-Art Methods}

\begin{figure}[t]
\centering
\includegraphics[width=\linewidth]{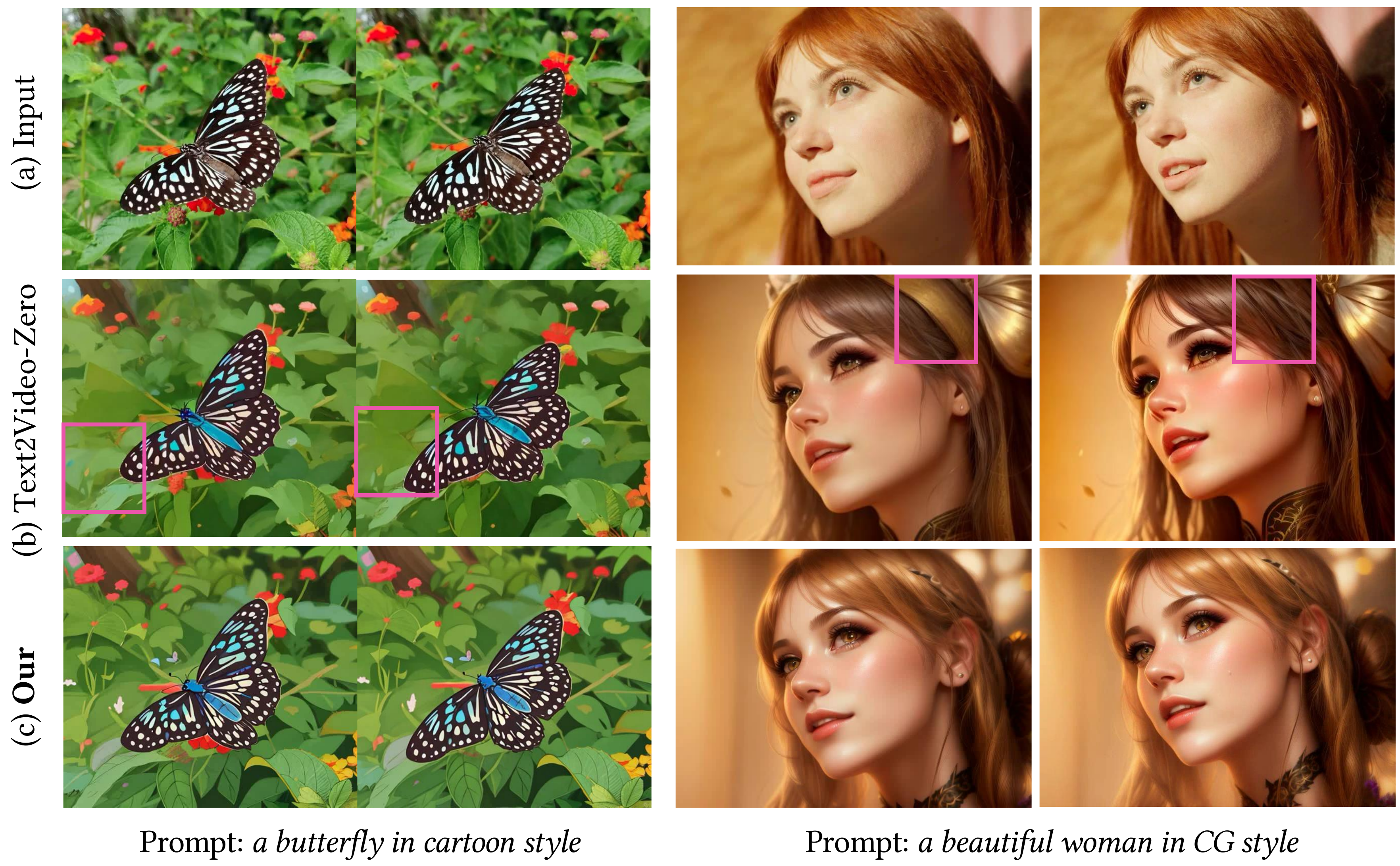}\vspace{1mm}
\includegraphics[width=\linewidth]{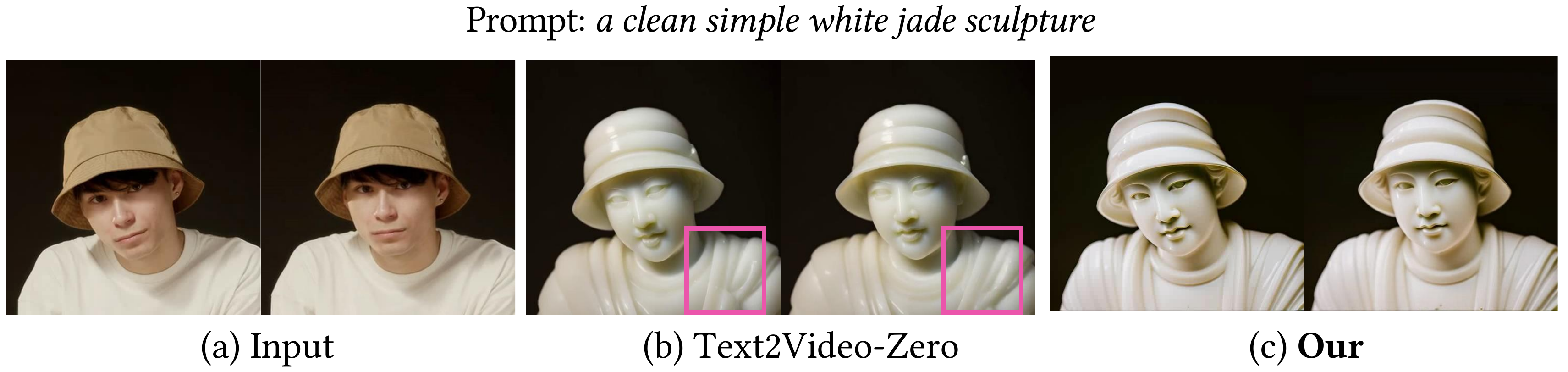}\vspace{-2mm}
\caption{Visual comparison with Text2Video-Zero. Text2Video-Zero and our method use the same customized model and ControlNet for a fair comparison. Our method outperforms Text2Video-Zero in terms of local texture temporal consistency. The red box indicates the inconsistent region.}\vspace{-4mm}
\label{fig:compare2}
\end{figure}

We compare with four recent zero-shot methods: vid2vid-zero~\cite{wang2023zero}, FateZero~\cite{qi2023fatezero}, Pix2Video~\cite{ceylan2023pix2video}, Text2Video-Zero~\cite{khachatryan2023text2video} on key frame translation with $K=5$. The official code of the first three methods does not support ControlNet, and when loading customized models, we find they fail to generate plausible results, \eg, vid2vid-zero will generate frames totally different from the input. Therefore, only Text2Video-Zero and our method use the customized model with ControlNet.
Figure~\ref{fig:compare} and Figure~\ref{fig:compare2} present the visual results. FateZero successfully reconstructs the input frame but fails to adjust it to match the prompt. On the other hand, vid2vid-zero and Pix2Video excessively modify the input frame, leading to significant shape distortion and discontinuity across frames.
While each frame generated by Text2Video-Zero exhibits high quality, they lack coherence in local textures as indicated by the black boxes. Finally, our proposed method demonstrates clear superiority in terms of output quality, content and prompt matching and temporal consistency.

For quantitative evaluation, we follow FateZero and Pix2Video to report
Fram-Acc (CLIP-based frame-wise editing accuracy),
Tmp-Con (CLIP-based cosine similarity between consecutive frames),
Pixel-MSE (averaged mean-squared pixel error between aligned consecutive frames) in Table~\ref{tb:quantitative_evaluation}. Our method achieves the best temporal consistency and the second best frame editing accuracy.
We further conduct a user study with 30 participants. The participants are asked to select the best results among the five methods based on three criteria: 1) how well the result balance between the prompt and the input frame, 2) the temporal consistency of the result, and 3) the overall quality of the video translation. Table~\ref{tb:quantitative_evaluation} presents the average preference rates across 8 testing videos, and our method achieves the highest rates in all three metrics.

\begin{table}[]
\begin{center}
\caption{Quantitative comparison and user preference rates.}\vspace{-2mm}
\resizebox{\linewidth}{!}{
\begin{tabular}{l|ccccc}
\toprule
\textbf{Metric}  & v2v-zero & FateZero & Pix2Video & T2V-Zero & Ours\\
\midrule
Fram-Acc & 0.862 & 0.556 & \textbf{0.995} & 0.963 & 0.979 \\
Tem-Con & 0.975 & 0.979 & 0.953 & \textbf{0.983} & \textbf{0.983} \\
Pixel-MSE & 0.098 & 0.085 & 0.216 & 0.084 & \textbf{0.069} \\
\midrule
User-Balance & 3.8\% & 5.9\% & 9.2\% & 15.4\% & \textbf{65.8\%} \\
User-Temporal & 3.8\% & 9.6\% & 4.2\% & 10.8\% & \textbf{71.6\%} \\
User-Overall & 2.9\% & 4.2\% & 4.2\% & 15.0\% & \textbf{73.7\%}\\
\bottomrule
\end{tabular}}\vspace{-4mm}
\label{tb:quantitative_evaluation}
\end{center}
\end{table}

\begin{figure*}[htbp]
\centering
\includegraphics[width=\linewidth]{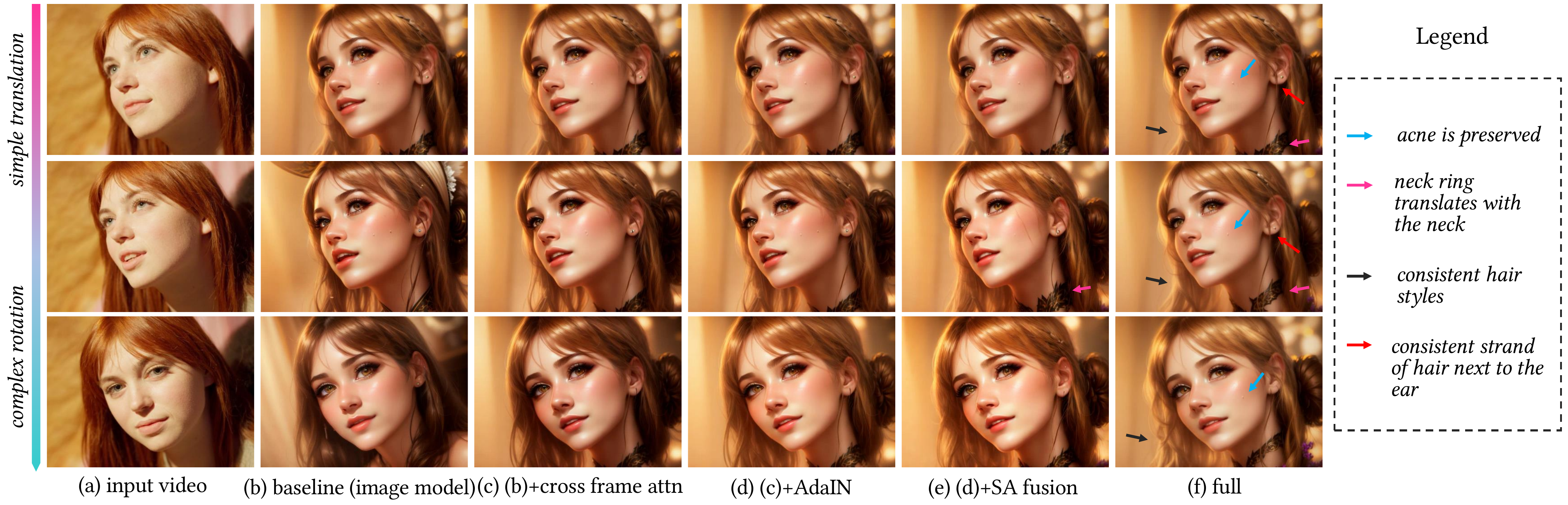}\vspace{-2mm}
\caption{Effect of the proposed hierarchical cross-frame constraints. (a) Input frames \#1, \#55, \#94. (b) Image diffusion model renders each frame independently. (c) Cross frame attention keeps the overall style consistent. (d) AdaIN preserves the hair color. (e) Shape-aware latent fusion keeps the overall movement of the objects coherent. (f) Pixel-aware latent fusion achieves pixel-level temporal consistency.}\vspace{-2mm}
\label{fig:ablation1}
\end{figure*}

\subsection{Ablation Study}\vspace{-3mm}

\noindent
\paragraph{Hierarchical cross-frame consistency constraints} Figure~\ref{fig:ablation1} compares the results with and without different cross-frame consistency constraints. We demonstrate the efficacy of our approach on a video containing simple translational motion in the first half and complex 3D rotation transformations in the latter half. To better evaluate the temporal consistency, we encourage readers to watch the videos on the project webpage.
The cross-frame attention ensures consistency in global style, while the adaptive latent adjustment in Sec.~\ref{sec:method1-4} maintains the same hair color as the first frame, or the hair color will follow the input frame to turn dark. Note that the adaptive latent adjustment is optional to allow users to decide which color to follow.
The above two global constraints cannot capture local movement. The shape-aware latent fusion (SA fusion) in Sec.~\ref{sec:method1-2} addresses this by translating the latent features to translate the neck ring, but cannot maintain pixel-level consistency for complex motion.
Only the proposed pixel-aware latent fusion (PA fusion) can coherently render local details such as hair styles and acne.

\begin{figure}[t]
\centering
\includegraphics[width=\linewidth]{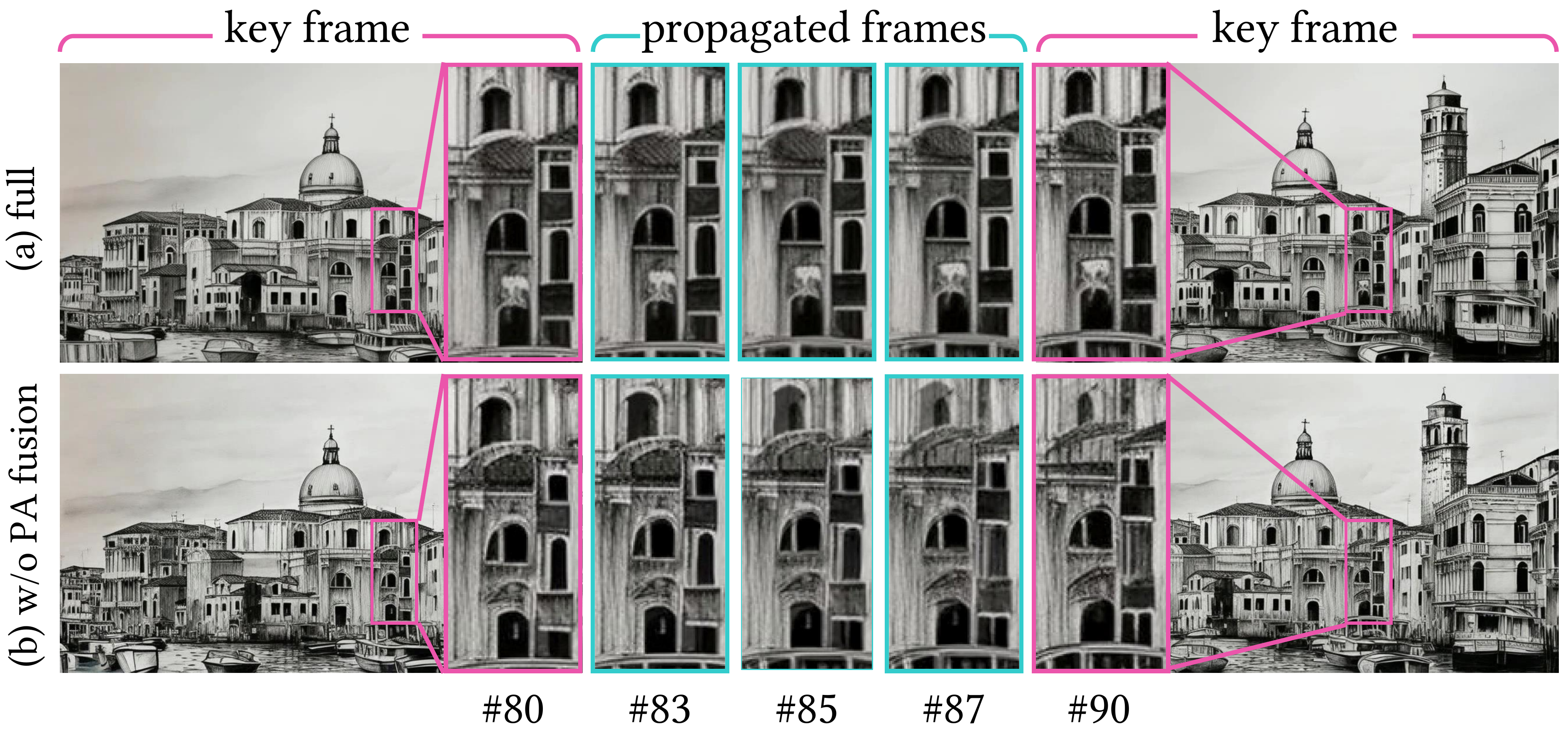}\vspace{-2mm}
\caption{Effect of the pixel-aware latent fusion on frame propagation. The proposed pixel-aware latent fusion helps generate consistent key frames. Without it, the pixel level inconsistency between key frames leads to ghosting artifacts on the non-key frames during the frame blending.}\vspace{-2mm}
\label{fig:ablation-pafusion2}
\end{figure}

\begin{figure}[t]
\centering
\includegraphics[width=\linewidth]{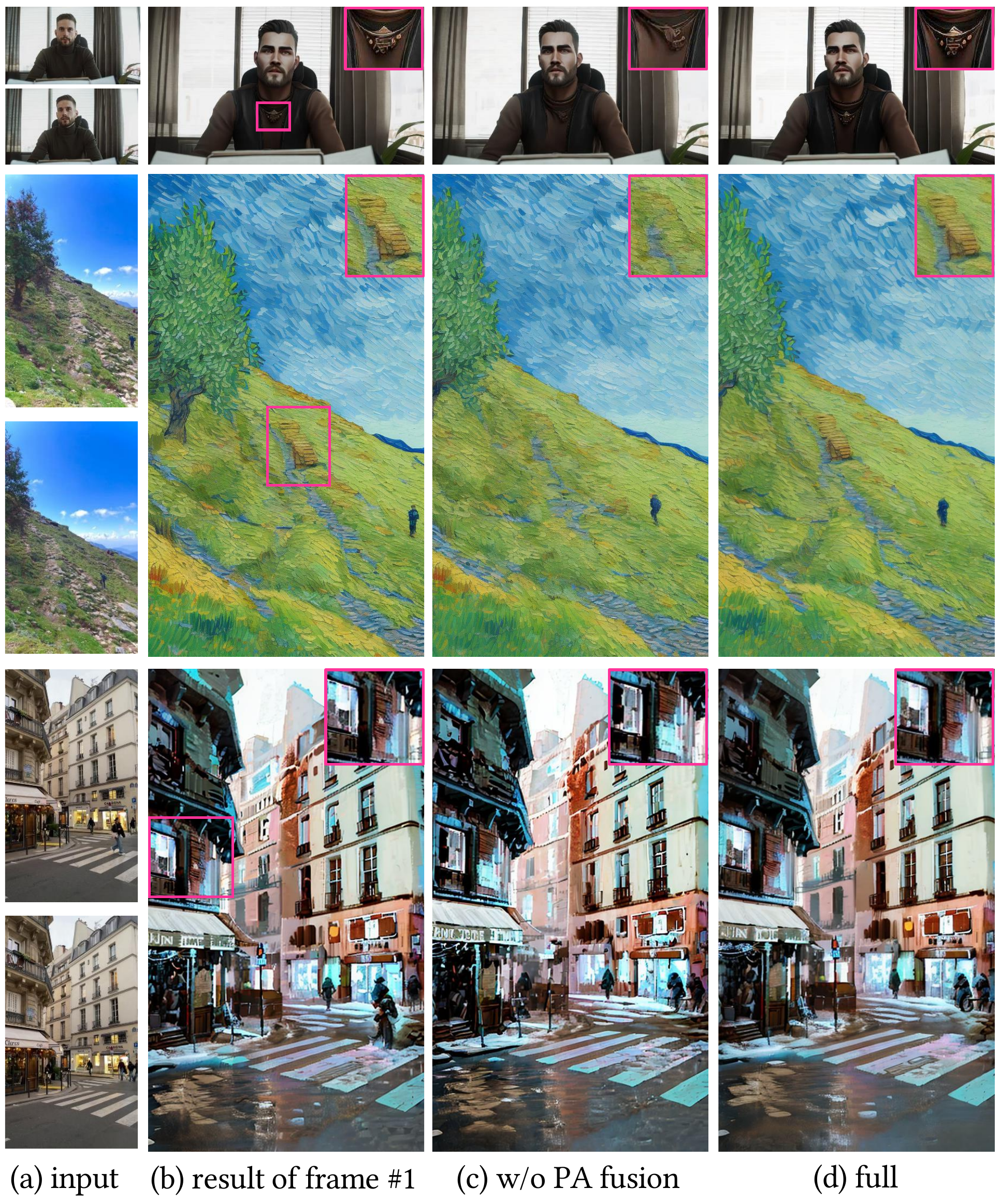}\vspace{-2mm}
\caption{Effect of the pixel-aware latent fusion. Prompts (from top to bottom): `\textit{Arcane style, a handsome man}', `\textit{Loving Vincent, hiking, grass}', `\textit{Disco Elysium, street view}'. Local regions are enlarged and shown in the top right.}\vspace{-2mm}
\label{fig:ablation-pafusion}
\end{figure}

We provide additional examples in Figs.~\ref{fig:ablation-pafusion2}-\ref{fig:ablation-pafusion} to demonstrate the effectiveness of PA fusion. While ControlNet can guide the structure well, the inherent randomness introduced by noise addition and denoising makes it difficult to maintain coherence in local textures, resulting in missing elements and altered details. The proposed PA fusion restores these details by utilizing the corresponding pixel information from previous frames. Moreover, such consistency between key frames can effectively reduce the ghosting artifacts in interpolated non key frames.

\begin{figure}[htbp]
\centering
\includegraphics[width=\linewidth]{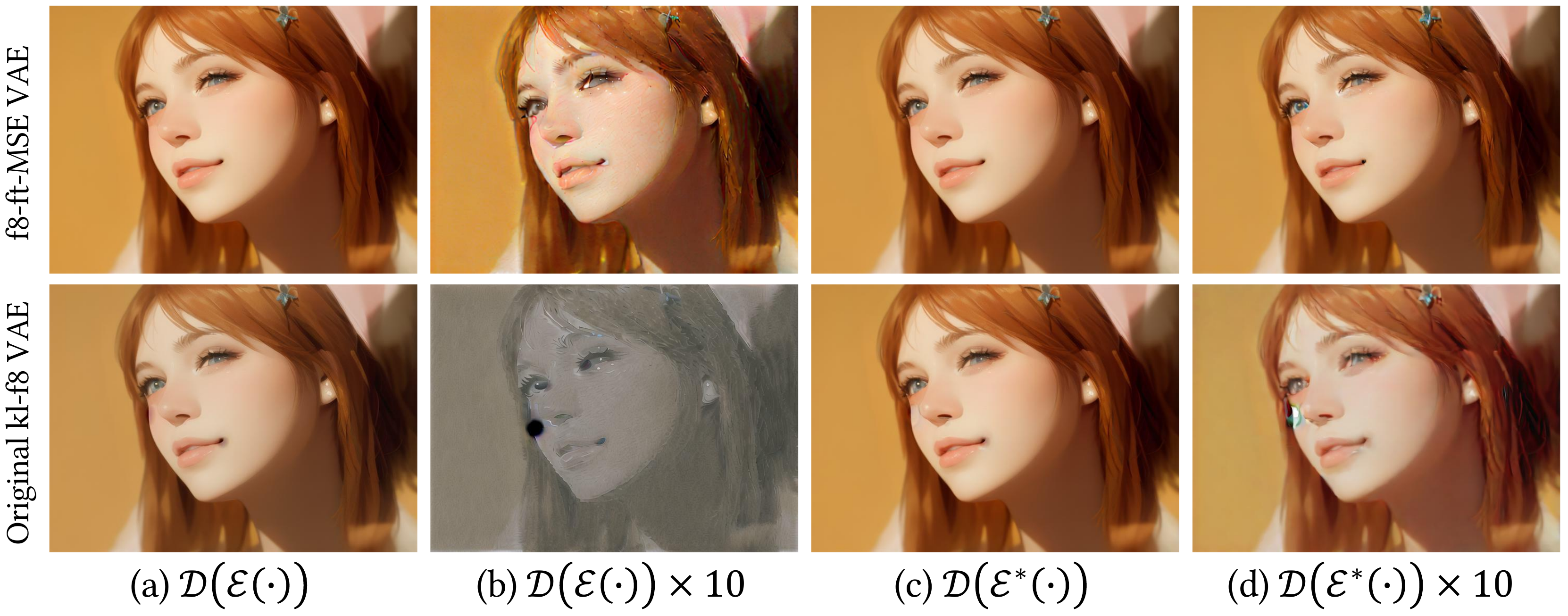}\vspace{-2mm}
\caption{The fidelity-oriented image encoding on two VAEs.}\vspace{-2mm}
\label{fig:ablation2}
\end{figure}
\begin{figure}[htbp]
\centering
\includegraphics[width=\linewidth]{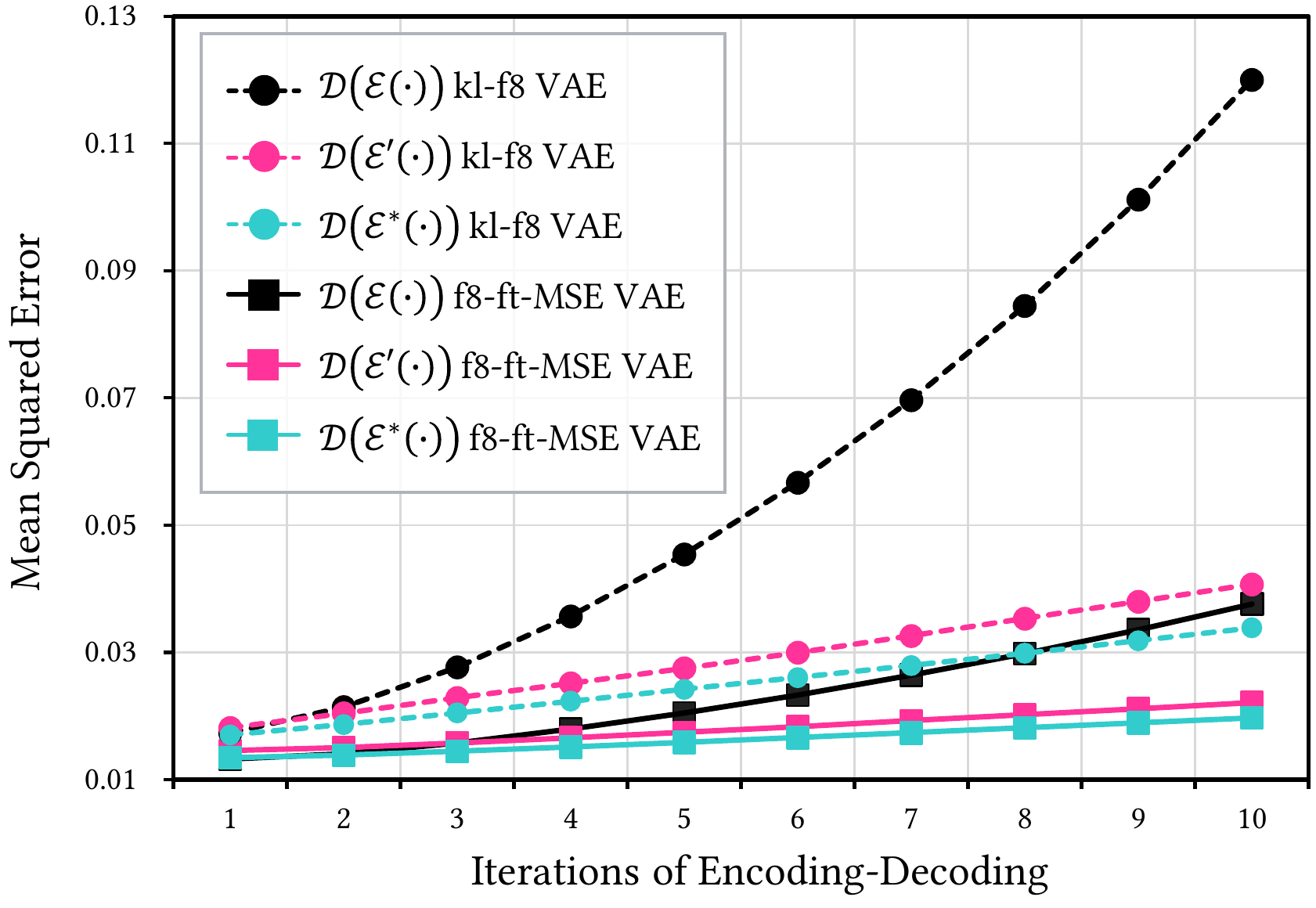}\vspace{-3mm}
\caption{Quantitative evaluation of image encoding schemes.}\vspace{-4mm}
\label{fig:ablation2_quantitative}
\end{figure}

\begin{figure}[t]
\centering
\includegraphics[width=\linewidth]{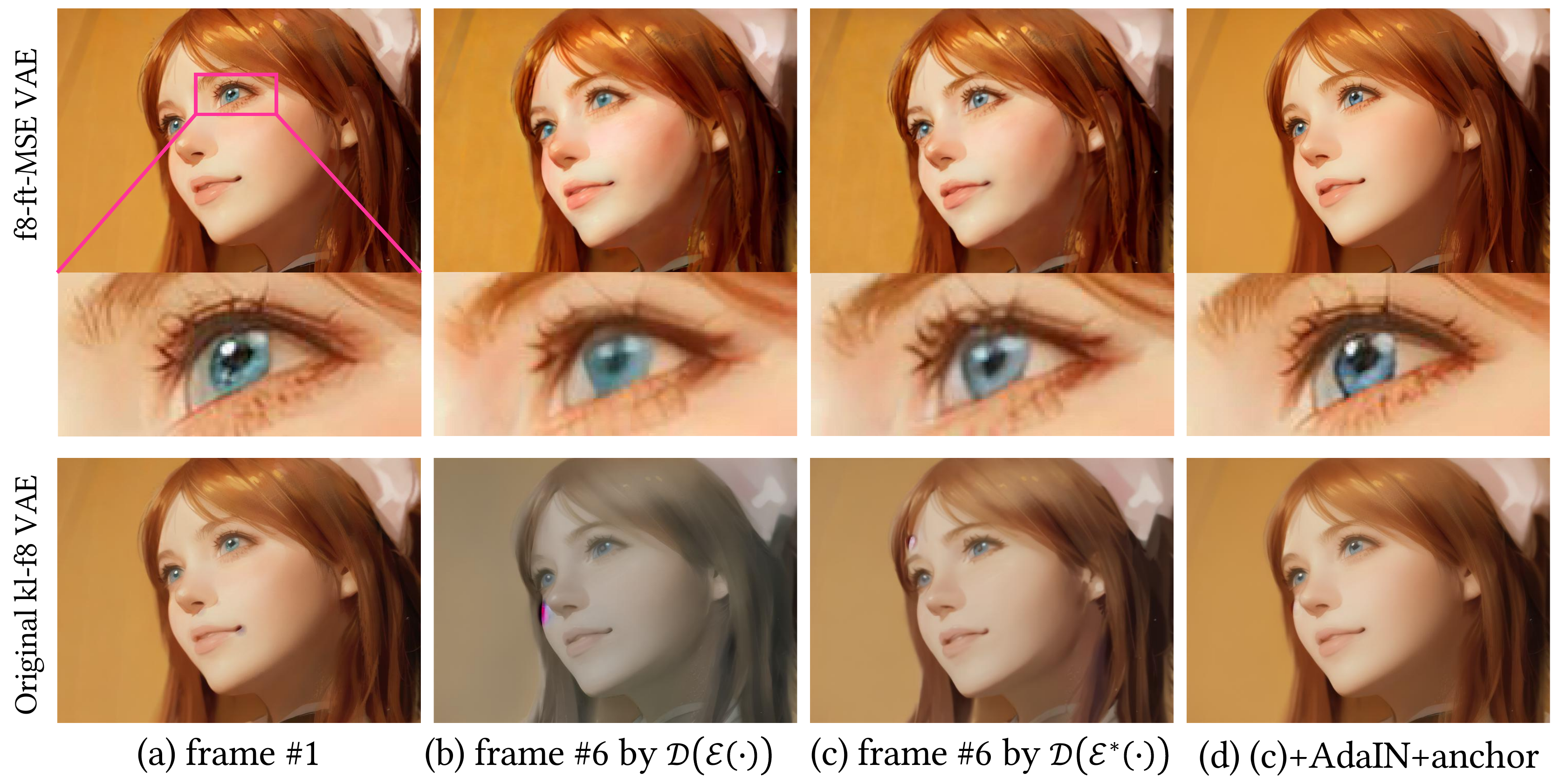}\vspace{-2mm}
\caption{Different constraints to prevent error accumulation.}\vspace{-2mm}
\label{fig:ablation3_encoder}
\end{figure}

\noindent
\paragraph{Fidelity-oriented image encoding}
We present a detailed analysis of our fidelity-oriented image encoding in Figs.~\ref{fig:ablation2}-\ref{fig:ablation3_encoder}, in addition to Fig.~\ref{fig:encoding}. Two Stable Diffusion's officially released autoencoders, the fine-tuned f8-ft-MSE VAE and the original more lossy kl-f8 VAE, are used for testing our method.
The fine-tuned VAE introduces artifacts and the original VAE results in great color bias as in Fig.~\ref{fig:ablation2}(b). Our proposed fidelity-oriented image encoding effectively alleviates these issues.
For quantitative evaluation, we report the MSE between the input image and the reconstructed result after multiple encoding and decoding in Fig.~\ref{fig:ablation2_quantitative}, using the first 1,000 images of the MS-COCO~\cite{lin2014microsoft} validation set. The results are consistent with the visual observations: our proposed method significantly reduces error accumulation compared to raw encoding methods.
Finally, we validate our encoding method in the video translation process in Fig.~\ref{fig:ablation3_encoder}(b)(c), where we use only the previous frame without the anchor frame in Eq.~(\ref{eq:warp}) to better visualize error accumulation. Our method mostly reduces the loss of details and color bias caused by lossy encoding. Besides, our pipeline includes an anchor frame and adaptive latent adjustment to further regulate the translation,
as shown in Fig.~\ref{fig:ablation3_encoder}(d), where no obvious errors are observed.

\paragraph{Frequency of key frames $K$} We report the quantitative full video translation results of Fig.~\ref{fig:ablation1}(a) under different $K$ in Table~\ref{tb:quantitative_evaluation2}. With large $K$, more frame interpolation improves pixel-level temporal consistency, which however harms the quality, leading to low Fram-Acc. A broad range of $K\in[5,20]$ is recommended for balance.

\begin{table}[]
\begin{center}
\caption{Effect of key frame sampling interval $K$}\vspace{-3mm}
\resizebox{0.9\linewidth}{!}{
\begin{tabular}{l|cccccc}
\toprule
\textbf{Metric}  & $K=1$&$K=5$&$K=10$&$K=20$&$K=50$ &$K=100$\\
\midrule
Fram-Acc & \textbf{1.000}&\textbf{1.000}&\textbf{1.000}&\textbf{1.000}&0.990&0.890\\
Tem-Con & 0.992 & 0.993 & \textbf{0.994} & \textbf{0.994} & 0.993 & 0.993 \\
Pixel-MSE & 0.037 & 0.028 & 0.025 & 0.022 & \textbf{0.020} & \textbf{0.020} \\
\bottomrule
\end{tabular}}\vspace{-2mm}
\label{tb:quantitative_evaluation2}
\end{center}
\end{table}

\begin{figure}[t]
\centering
\includegraphics[width=0.98\linewidth]{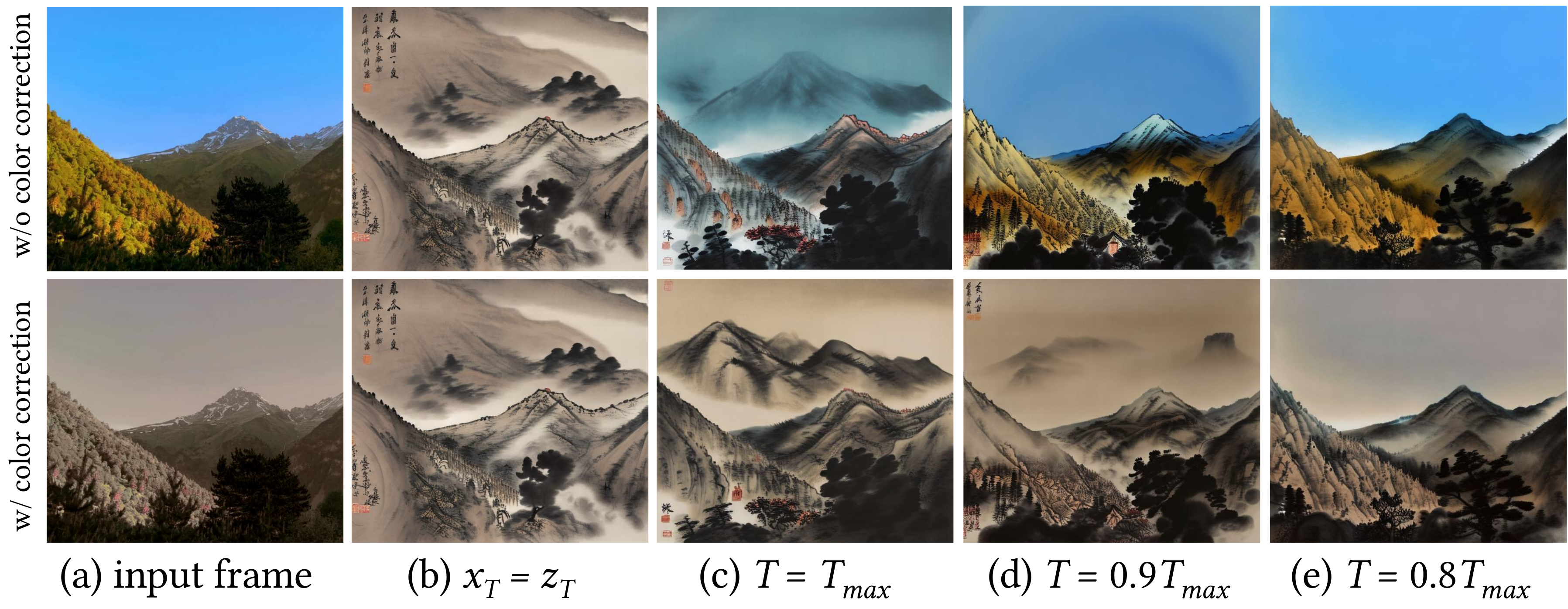}\vspace{-2mm}
\caption{Effect of the initialization of $x_T$. Prompt: a traditional mountain in Chinese ink wash painting. The proposed framework enables flexible content and color control by adjusting $T$ and color correction.}\vspace{-2mm}
\label{fig:ablation_init}
\end{figure}

\subsection{More Results}\vspace{-3mm}

\noindent
\paragraph{Flexible structure and color control} The proposed pipeline allows flexible control over content preservation through the initialization of $x_T$. Rather than setting $x_T$ to a Gaussian noise (Fig.~\ref{fig:ablation_init}(b)), we use a noisy latent version of the input frame to better preserve details (Fig.~\ref{fig:ablation_init}(c)). Users can adjust the value of $T$ to balance content and prompt.
Moreover, if the input frame introduces unwanted color bias (\eg, blue sky in Chinese ink painting), a color correction option is provided: the input frame is adjusted to match the color histogram of the frame generated by $x_T=z_T$ (Fig.~\ref{fig:ablation_init}(b)). With the adjusted frame as input (bottom row of Fig.~\ref{fig:ablation_init}(a)), the rendered results (bottom row of Figs.~\ref{fig:ablation_init}(c)-(f)) better match the color indicated by the prompt.

\begin{figure}[t]
\centering
\includegraphics[width=0.98\linewidth]{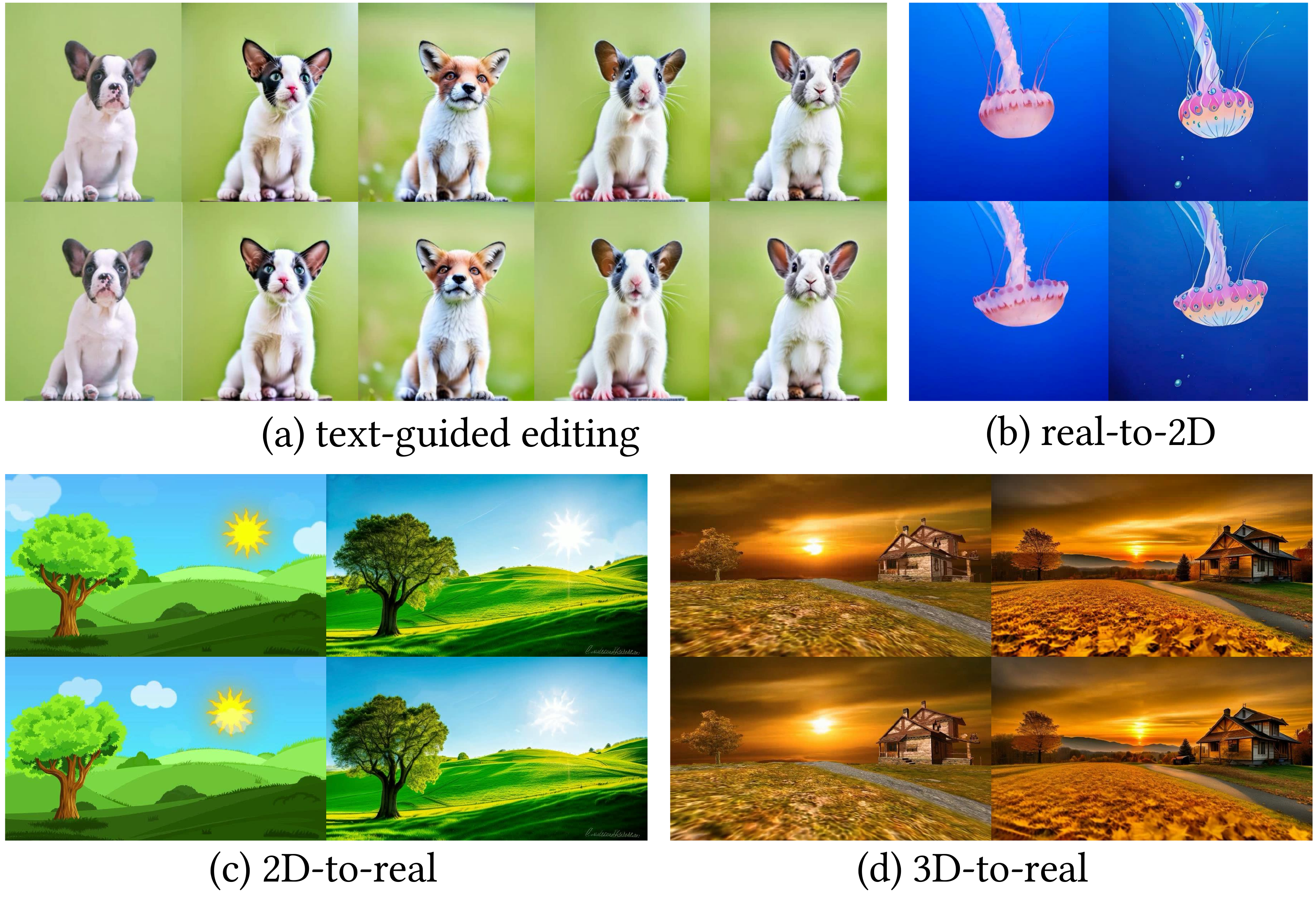}\vspace{-3mm}
\caption{Applications of the proposed method.}\vspace{-2mm}
\label{fig:app}
\end{figure}

\begin{figure}[t]
\centering
\includegraphics[width=0.98\linewidth]{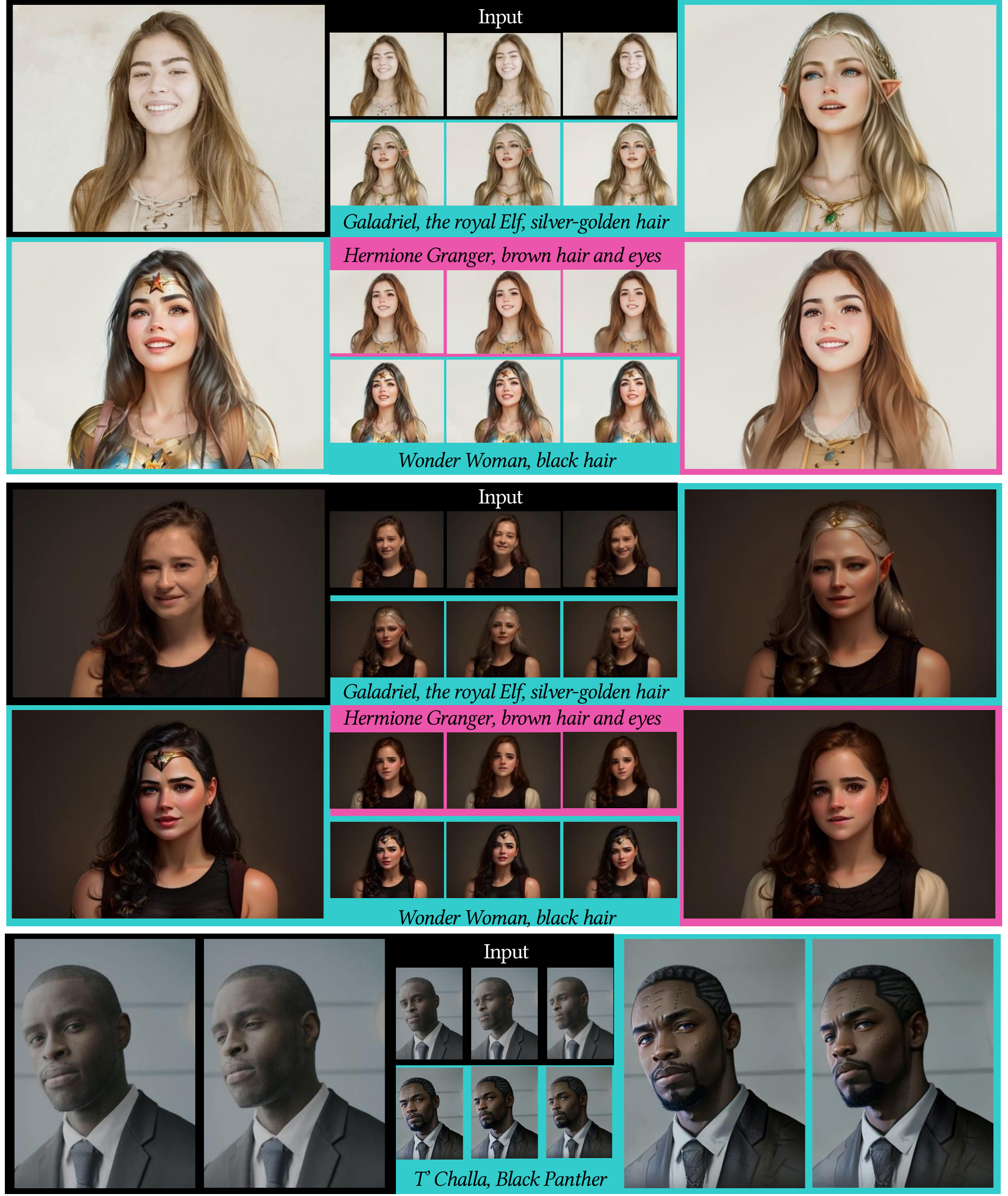}\vspace{-2mm}
\caption{Applications: text-guided virtual character generation. Results are generated with a single image diffusion model.}\vspace{-4mm}
\label{fig:app2}
\end{figure}

\begin{figure}[t]
\centering
\includegraphics[width=\linewidth]{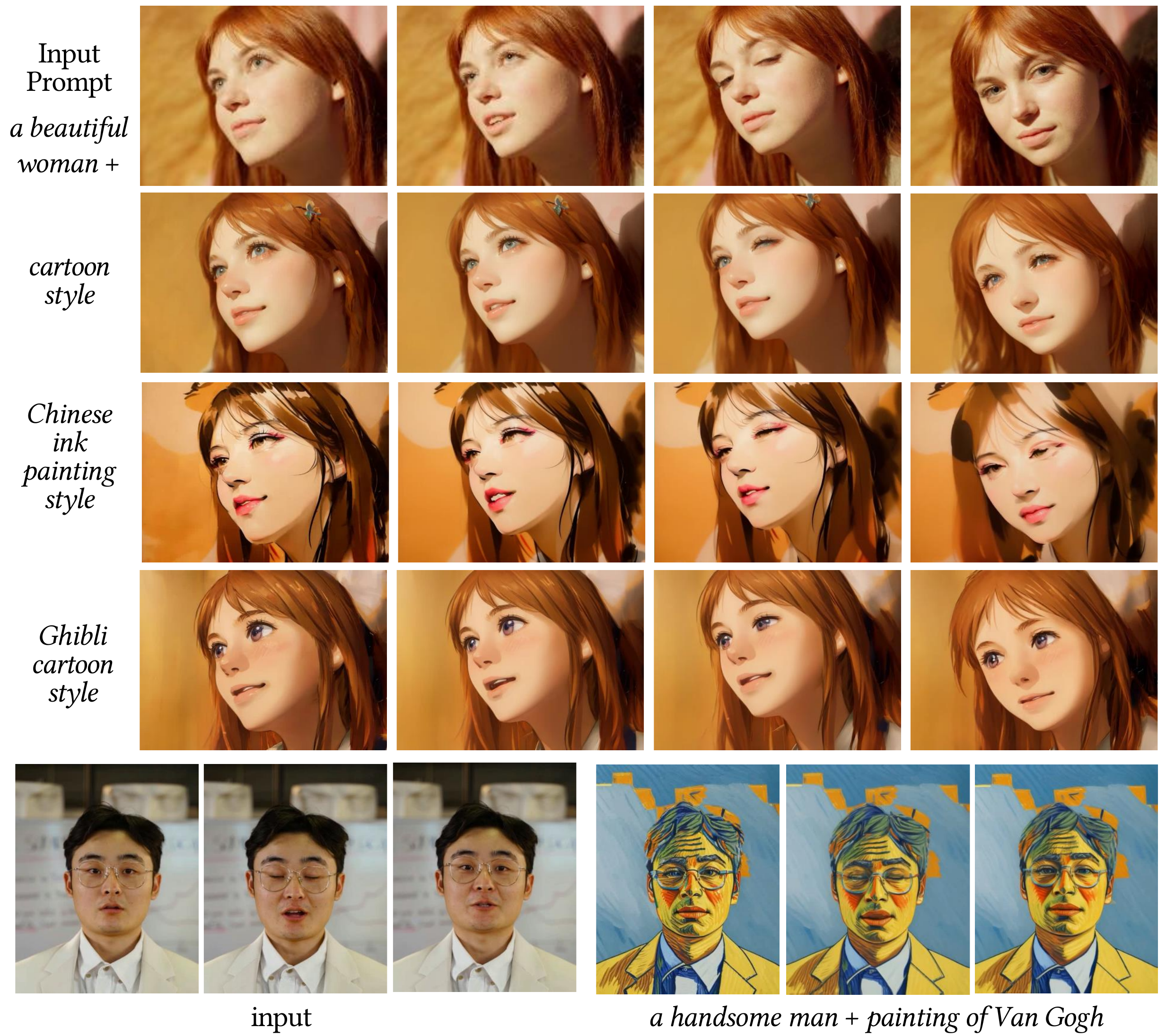}\vspace{-3mm}
\caption{Applications: video stylization. Thanks to the compatible design, our method can use off-the-shelf pre-trained image models customized for different styles to accurately stylize videos.}\vspace{-2mm}
\label{fig:more_result}
\end{figure}

\noindent
\paragraph{Applications} Figure~\ref{fig:app} shows some applications of our method. With prompts `a cute cat/fox/hamster/rabbit', we can perform text-guided editing to translate a dog into other kinds of pets in Fig.~\ref{fig:app}(a). By using customized modes for generating cartoons or photos, we can achieve non-photorealistic and photorealistic rendering in Fig.~\ref{fig:app}(b) and Figs.~\ref{fig:app}(c)(d), respectively.
In Fig.~\ref{fig:app2}, we present our synthesized dynamic virtual characters of novels and manga, based on a real human video and a prompt to describe the appearance. Additional results are shown in Fig.~\ref{fig:more_result}.

\subsection{Limitations}

\begin{figure}[t]
\centering
\includegraphics[width=\linewidth]{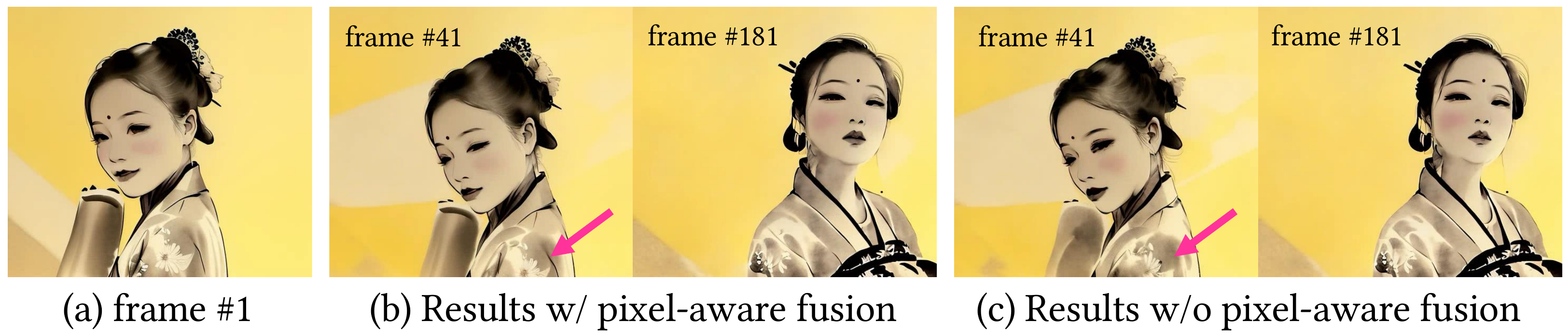}\vspace{-2mm}
\caption{Limitation: failure optical flow due to large motions. Our method is not suitable for processing videos where it is difficult to estimate the optical flow.}\vspace{-3mm}
\label{fig:limitation0}
\end{figure}

\begin{figure}[t]
\centering
\includegraphics[width=\linewidth]{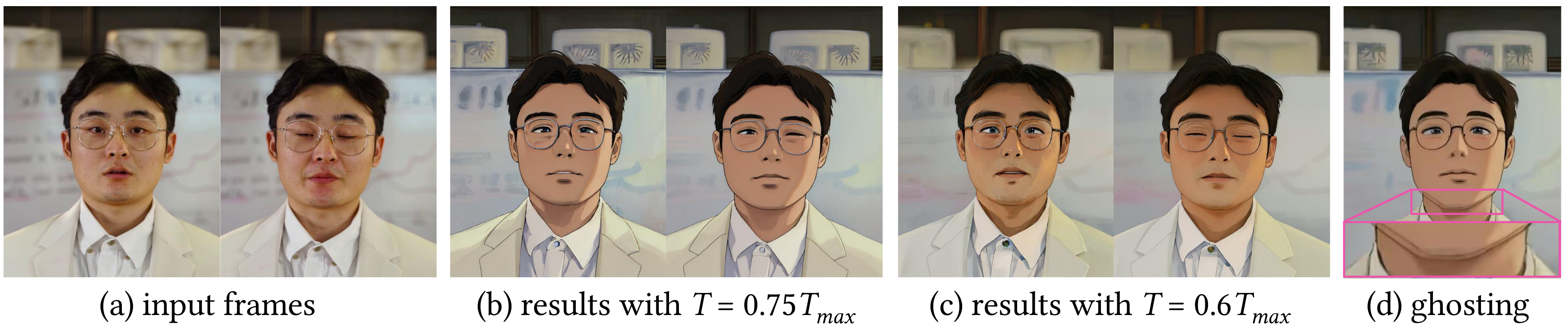}\vspace{-2mm}
\caption{Limitation: trade-off between content and prompt. 
}\vspace{-2mm}
\label{fig:limitation1}
\end{figure}

\begin{figure}[t]
\centering
\includegraphics[width=\linewidth]{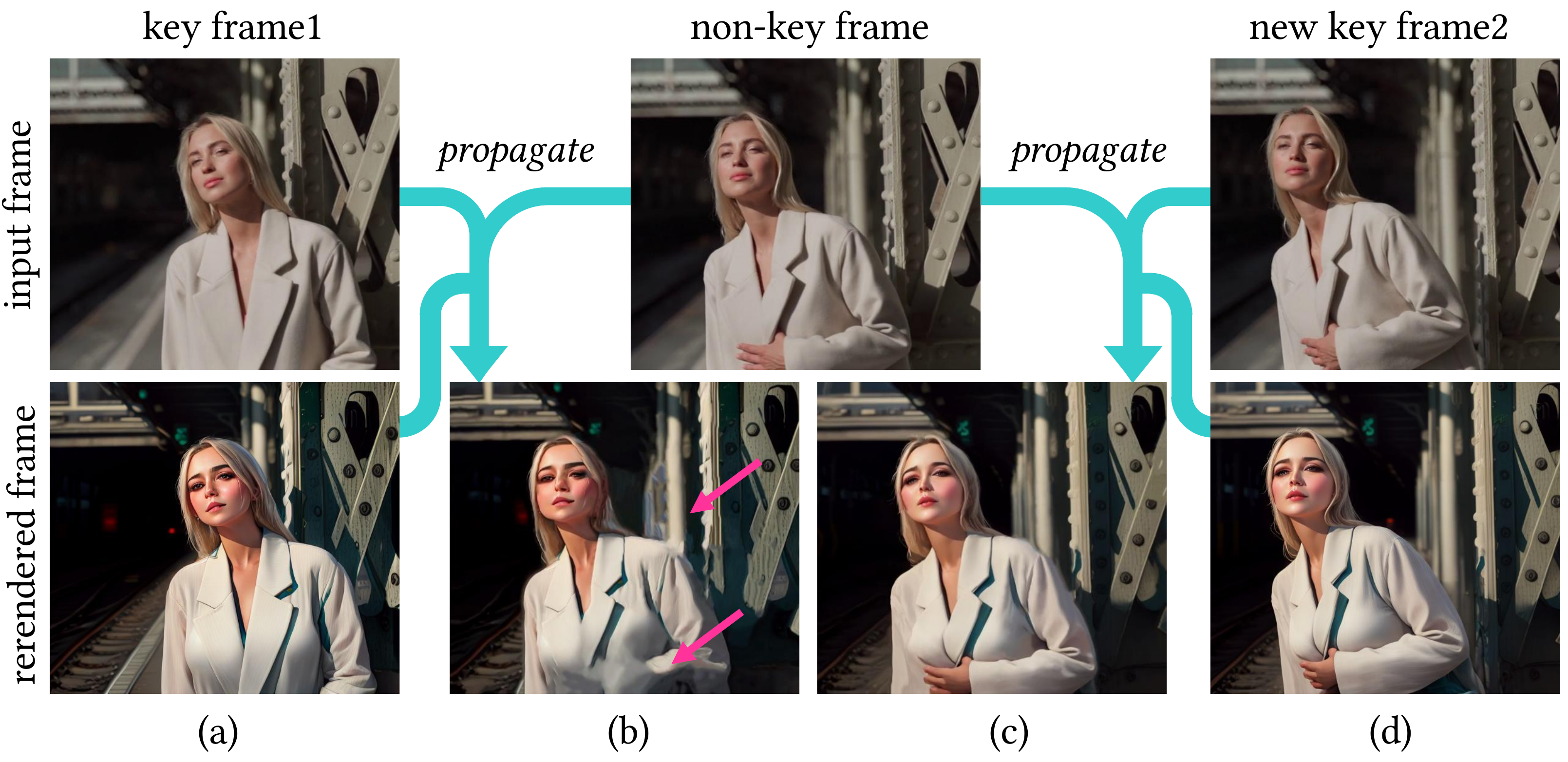}\vspace{-3mm}
\caption{Limitation: failed propagation w/o good key frames.}\vspace{-3mm}
\label{fig:limitation2}
\end{figure}

Figures~\ref{fig:limitation0}-\ref{fig:limitation2} illustrate typical failure cases of our method. First, our method relies on optical flow and therefore, inaccurate optical flow can lead to artifacts. In Fig.~\ref{fig:limitation0}, our method can only preserve the embroidery if the cross-frame correspondence is available. Otherwise, the proposed PA fusion will have no effect.
Second, our method assumes the optical flow remains unchanged before and after translation, which may not hold true for significant appearance changes as in Fig.~\ref{fig:limitation1}(b), where the resulting movement may be wrong.
Although setting a smaller $T$ can address this issue, it may compromise the desired styles. Meanwhile, the mismatches of the optical flow mean the mismatches in the translated key frames, which may lead to ghosting artifacts (Fig.~\ref{fig:limitation1}(d)) after temporal-aware blending. Also, we find that small details and subtle motions like accessories and eye movement cannot be well preserved during the translation.
Lastly, we uniformly sample the key frames, which may not optimal. Ideally, the key frames should contain all unique objects; otherwise, the propagation cannot create unseen content such as the hand in Fig.~\ref{fig:limitation2}(b). One potential solution is user-interactive translation, where users can manually assign new key frames based on the previous results.

\section{Conclusion}

This paper presents a zero-shot framework to adapt image diffusion models for video translation. Our method utilizes hierarchical cross-frame constraints to enforce temporal consistency in both global style and low-level textures, leveraging the key optical flow. The compatibility with existing image diffusion techniques indicates that our idea might be applied to other text-guided video editing tasks, such as video super-resolution and inpainting. Additionally, our proposed fidelity-oriented image encoding could benefit existing diffusion-based methods. We believe that our approach can facilitate the creation of high-quality and temporally-coherent videos and inspire further research in this field.

\noindent
\textbf{Acknowledgments.} This study is supported under the RIE2020 Industry Alignment Fund Industry Collaboration Projects (IAF-ICP) Funding Initiative, as well as cash and in-kind contribution from the industry partner(s). It is also supported by Singapore MOE AcRF Tier 2 (MOE-T2EP20221-0011, MOE-T2EP20221-0012) and NTU NAP.

{\small
\bibliographystyle{ieee_fullname}
\bibliography{sample-bibliography}

\begin{thebibliography}{10}\itemsep=-1pt

\bibitem{avrahami2022blended}
Omri Avrahami, Ohad Fried, and Dani Lischinski.
\newblock Blended latent diffusion.
\newblock {\em arXiv preprint arXiv:2206.02779}, 2022.

\bibitem{brooks2022instructpix2pix}
Tim Brooks, Aleksander Holynski, and Alexei~A Efros.
\newblock {InstructPix2Pix}: Learning to follow image editing instructions.
\newblock {\em arXiv preprint arXiv:2211.09800}, 2022.

\bibitem{ceylan2023pix2video}
Duygu Ceylan, Chun-Hao~Paul Huang, and Niloy~J Mitra.
\newblock Pix2video: Video editing using image diffusion.
\newblock {\em arXiv preprint arXiv:2303.12688}, 2023.

\bibitem{croitoru2023diffusion}
Florinel-Alin Croitoru, Vlad Hondru, Radu~Tudor Ionescu, and Mubarak Shah.
\newblock Diffusion models in vision: A survey.
\newblock {\em {IEEE} Transactions on Pattern Analysis and Machine
  Intelligence}, 2023.

\bibitem{darabi2012image}
Soheil Darabi, Eli Shechtman, Connelly Barnes, Dan~B Goldman, and Pradeep Sen.
\newblock Image melding: Combining inconsistent images using patch-based
  synthesis.
\newblock {\em {ACM} Transactions on Graphics}, 31(4):82--1, 2012.

\bibitem{ding2021cogview}
Ming Ding, Zhuoyi Yang, Wenyi Hong, Wendi Zheng, Chang Zhou, Da Yin, Junyang
  Lin, Xu Zou, Zhou Shao, Hongxia Yang, et~al.
\newblock Cogview: Mastering text-to-image generation via transformers.
\newblock In {\em Advances in Neural Information Processing Systems},
  volume~34, pages 19822--19835, 2021.

\bibitem{esser2021taming}
Patrick Esser, Robin Rombach, and Bjorn Ommer.
\newblock Taming transformers for high-resolution image synthesis.
\newblock In {\em Proc.~IEEE Int'l Conf.~Computer Vision and Pattern
  Recognition}, pages 12873--12883, 2021.

\bibitem{fivser2017example}
Jakub Fi{\v{s}}er, Ond{\v{r}}ej Jamri{\v{s}}ka, David Simons, Eli Shechtman,
  Jingwan Lu, Paul Asente, Michal Luk{\'a}{\v{c}}, and Daniel S{\`y}kora.
\newblock Example-based synthesis of stylized facial animations.
\newblock {\em ACM Transactions on Graphics (TOG)}, 36(4):1--11, 2017.

\bibitem{gafni2022make}
Oran Gafni, Adam Polyak, Oron Ashual, Shelly Sheynin, Devi Parikh, and Yaniv
  Taigman.
\newblock Make-a-scene: Scene-based text-to-image generation with human priors.
\newblock In {\em Proc.~European Conf.~Computer Vision}, pages 89--106.
  Springer, 2022.

\bibitem{gal2022image}
Rinon Gal, Yuval Alaluf, Yuval Atzmon, Or Patashnik, Amit~H Bermano, Gal
  Chechik, and Daniel Cohen-Or.
\newblock An image is worth one word: Personalizing text-to-image generation
  using textual inversion.
\newblock {\em arXiv preprint arXiv:2208.01618}, 2022.

\bibitem{heitz2018high}
Eric Heitz and Fabrice Neyret.
\newblock High-performance by-example noise using a histogram-preserving
  blending operator.
\newblock {\em Proceedings of the ACM on Computer Graphics and Interactive
  Techniques}, 1(2):1--25, 2018.

\bibitem{hertz2022prompt}
Amir Hertz, Ron Mokady, Jay Tenenbaum, Kfir Aberman, Yael Pritch, and Daniel
  Cohen-Or.
\newblock Prompt-to-prompt image editing with cross attention control.
\newblock {\em arXiv preprint arXiv:2208.01626}, 2022.

\bibitem{Hertzmann2001Image}
Aaron Hertzmann, Charles~E. Jacobs, Nuria Oliver, Brian Curless, and David~H.
  Salesin.
\newblock Image analogies.
\newblock In {\em Proc. Conf.~Computer Graphics and Interactive Techniques},
  pages 327--340, 2001.

\bibitem{ho2022imagen}
Jonathan Ho, William Chan, Chitwan Saharia, Jay Whang, Ruiqi Gao, Alexey
  Gritsenko, Diederik~P Kingma, Ben Poole, Mohammad Norouzi, David~J Fleet,
  et~al.
\newblock Imagen video: High definition video generation with diffusion models.
\newblock {\em arXiv preprint arXiv:2210.02303}, 2022.

\bibitem{ho2020denoising}
Jonathan Ho, Ajay Jain, and Pieter Abbeel.
\newblock Denoising diffusion probabilistic models.
\newblock In {\em Advances in Neural Information Processing Systems},
  volume~33, pages 6840--6851, 2020.

\bibitem{hovideo}
Jonathan Ho, Tim Salimans, Alexey~A Gritsenko, William Chan, Mohammad Norouzi,
  and David~J Fleet.
\newblock Video diffusion models.
\newblock In {\em Advances in Neural Information Processing Systems}, 2022.

\bibitem{hulora}
Edward~J Hu, Phillip Wallis, Zeyuan Allen-Zhu, Yuanzhi Li, Shean Wang, Lu Wang,
  Weizhu Chen, et~al.
\newblock Lora: Low-rank adaptation of large language models.
\newblock In {\em Proc.~Int'l Conf.~Learning Representations}, 2021.

\bibitem{huang2017adain}
Xun Huang and Serge Belongie.
\newblock Arbitrary style transfer in real-time with adaptive instance
  normalization.
\newblock In {\em Proc.~Int'l Conf.~Computer Vision}, pages 1510--1519, 2017.

\bibitem{jamrivska2019stylizing}
Ond{\v{r}}ej Jamri{\v{s}}ka, {\v{S}}{\'a}rka Sochorov{\'a}, Ond{\v{r}}ej
  Texler, Michal Luk{\'a}{\v{c}}, Jakub Fi{\v{s}}er, Jingwan Lu, Eli Shechtman,
  and Daniel S{\`y}kora.
\newblock Stylizing video by example.
\newblock {\em {ACM} Transactions on Graphics}, 38(4):1--11, 2019.

\bibitem{khachatryan2023text2video}
Levon Khachatryan, Andranik Movsisyan, Vahram Tadevosyan, Roberto Henschel,
  Zhangyang Wang, Shant Navasardyan, and Humphrey Shi.
\newblock Text2video-zero: Text-to-image diffusion models are zero-shot video
  generators.
\newblock {\em arXiv preprint arXiv:2303.13439}, 2023.

\bibitem{lin2014microsoft}
Tsung-Yi Lin, Michael Maire, Serge Belongie, James Hays, Pietro Perona, Deva
  Ramanan, Piotr Doll{\'a}r, and C~Lawrence Zitnick.
\newblock Microsoft coco: Common objects in context.
\newblock In {\em Proc.~European Conf.~Computer Vision}, pages 740--755.
  Springer, 2014.

\bibitem{liu2023video}
Shaoteng Liu, Yuechen Zhang, Wenbo Li, Zhe Lin, and Jiaya Jia.
\newblock Video-p2p: Video editing with cross-attention control.
\newblock {\em arXiv preprint arXiv:2303.04761}, 2023.

\bibitem{meng2021sdedit}
Chenlin Meng, Yutong He, Yang Song, Jiaming Song, Jiajun Wu, Jun-Yan Zhu, and
  Stefano Ermon.
\newblock Sdedit: Guided image synthesis and editing with stochastic
  differential equations.
\newblock In {\em Proc.~Int'l Conf.~Learning Representations}, 2021.

\bibitem{mokady2022null}
Ron Mokady, Amir Hertz, Kfir Aberman, Yael Pritch, and Daniel Cohen-Or.
\newblock Null-text inversion for editing real images using guided diffusion
  models.
\newblock {\em arXiv preprint arXiv:2211.09794}, 2022.

\bibitem{molad2023dreamix}
Eyal Molad, Eliahu Horwitz, Dani Valevski, Alex~Rav Acha, Yossi Matias, Yael
  Pritch, Yaniv Leviathan, and Yedid Hoshen.
\newblock Dreamix: Video diffusion models are general video editors.
\newblock {\em arXiv preprint arXiv:2302.01329}, 2023.

\bibitem{nichol2022glide}
Alexander~Quinn Nichol, Prafulla Dhariwal, Aditya Ramesh, Pranav Shyam, Pamela
  Mishkin, Bob Mcgrew, Ilya Sutskever, and Mark Chen.
\newblock Glide: Towards photorealistic image generation and editing with
  text-guided diffusion models.
\newblock In {\em Proc.~IEEE Int'l Conf.~Machine Learning}, pages 16784--16804,
  2022.

\bibitem{qi2023fatezero}
Chenyang Qi, Xiaodong Cun, Yong Zhang, Chenyang Lei, Xintao Wang, Ying Shan,
  and Qifeng Chen.
\newblock Fatezero: Fusing attentions for zero-shot text-based video editing.
\newblock {\em arXiv preprint arXiv:2303.09535}, 2023.

\bibitem{radford2021learning}
Alec Radford, Jong~Wook Kim, Chris Hallacy, Aditya Ramesh, Gabriel Goh,
  Sandhini Agarwal, Girish Sastry, Amanda Askell, Pamela Mishkin, Jack Clark,
  et~al.
\newblock Learning transferable visual models from natural language
  supervision.
\newblock In {\em Proc.~IEEE Int'l Conf.~Machine Learning}, pages 8748--8763.
  PMLR, 2021.

\bibitem{raffel2020exploring}
Colin Raffel, Noam Shazeer, Adam Roberts, Katherine Lee, Sharan Narang, Michael
  Matena, Yanqi Zhou, Wei Li, and Peter~J Liu.
\newblock Exploring the limits of transfer learning with a unified text-to-text
  transformer.
\newblock {\em The Journal of Machine Learning Research}, 21(1):5485--5551,
  2020.

\bibitem{ramesh2022hierarchical}
Aditya Ramesh, Prafulla Dhariwal, Alex Nichol, Casey Chu, and Mark Chen.
\newblock Hierarchical text-conditional image generation with clip latents.
\newblock {\em arXiv preprint arXiv:2204.06125}, 2022.

\bibitem{ramesh2021zero}
Aditya Ramesh, Mikhail Pavlov, Gabriel Goh, Scott Gray, Chelsea Voss, Alec
  Radford, Mark Chen, and Ilya Sutskever.
\newblock Zero-shot text-to-image generation.
\newblock In {\em Proc.~IEEE Int'l Conf.~Machine Learning}, pages 8821--8831,
  2021.

\bibitem{rombach2022high}
Robin Rombach, Andreas Blattmann, Dominik Lorenz, Patrick Esser, and Bj{\"o}rn
  Ommer.
\newblock High-resolution image synthesis with latent diffusion models.
\newblock In {\em Proc.~IEEE Int'l Conf.~Computer Vision and Pattern
  Recognition}, pages 10684--10695, 2022.

\bibitem{ruiz2022dreambooth}
Nataniel Ruiz, Yuanzhen Li, Varun Jampani, Yael Pritch, Michael Rubinstein, and
  Kfir Aberman.
\newblock Dreambooth: Fine tuning text-to-image diffusion models for
  subject-driven generation.
\newblock {\em arXiv preprint arXiv:2208.12242}, 2022.

\bibitem{saharia2022photorealistic}
Chitwan Saharia, William Chan, Saurabh Saxena, Lala Li, Jay Whang, Emily~L
  Denton, Kamyar Ghasemipour, Raphael Gontijo~Lopes, Burcu Karagol~Ayan, Tim
  Salimans, et~al.
\newblock Photorealistic text-to-image diffusion models with deep language
  understanding.
\newblock In {\em Advances in Neural Information Processing Systems},
  volume~35, pages 36479--36494, 2022.

\bibitem{shin2023edit}
Chaehun Shin, Heeseung Kim, Che~Hyun Lee, Sang-gil Lee, and Sungroh Yoon.
\newblock Edit-a-video: Single video editing with object-aware consistency.
\newblock {\em arXiv preprint arXiv:2303.07945}, 2023.

\bibitem{singer2022make}
Uriel Singer, Adam Polyak, Thomas Hayes, Xi Yin, Jie An, Songyang Zhang, Qiyuan
  Hu, Harry Yang, Oron Ashual, Oran Gafni, et~al.
\newblock Make-a-video: Text-to-video generation without text-video data.
\newblock In {\em Proc.~Int'l Conf.~Learning Representations}, 2023.

\bibitem{songdenoising}
Jiaming Song, Chenlin Meng, and Stefano Ermon.
\newblock Denoising diffusion implicit models.
\newblock In {\em Proc.~Int'l Conf.~Learning Representations}, 2021.

\bibitem{vaswani2017attention}
Ashish Vaswani, Noam Shazeer, Niki Parmar, Jakob Uszkoreit, Llion Jones,
  Aidan~N Gomez, {\L}ukasz Kaiser, and Illia Polosukhin.
\newblock Attention is all you need.
\newblock In {\em Advances in Neural Information Processing Systems},
  volume~30, 2017.

\bibitem{wang2023zero}
Wen Wang, Kangyang Xie, Zide Liu, Hao Chen, Yue Cao, Xinlong Wang, and Chunhua
  Shen.
\newblock Zero-shot video editing using off-the-shelf image diffusion models.
\newblock {\em arXiv preprint arXiv:2303.17599}, 2023.

\bibitem{wu2022tune}
Jay~Zhangjie Wu, Yixiao Ge, Xintao Wang, Weixian Lei, Yuchao Gu, Wynne Hsu,
  Ying Shan, Xiaohu Qie, and Mike~Zheng Shou.
\newblock Tune-a-video: One-shot tuning of image diffusion models for
  text-to-video generation.
\newblock {\em arXiv preprint arXiv:2212.11565}, 2022.

\bibitem{xu2022gmflow}
Haofei Xu, Jing Zhang, Jianfei Cai, Hamid Rezatofighi, and Dacheng Tao.
\newblock Gmflow: Learning optical flow via global matching.
\newblock In {\em Proc.~IEEE Int'l Conf.~Computer Vision and Pattern
  Recognition}, pages 8121--8130, 2022.

\bibitem{xu2018attngan}
Tao Xu, Pengchuan Zhang, Qiuyuan Huang, Han Zhang, Zhe Gan, Xiaolei Huang, and
  Xiaodong He.
\newblock Attngan: Fine-grained text to image generation with attentional
  generative adversarial networks.
\newblock In {\em Proc.~IEEE Int'l Conf.~Computer Vision and Pattern
  Recognition}, pages 1316--1324, 2018.

\bibitem{zhang2021cross}
Han Zhang, Jing~Yu Koh, Jason Baldridge, Honglak Lee, and Yinfei Yang.
\newblock Cross-modal contrastive learning for text-to-image generation.
\newblock In {\em Proc.~IEEE Int'l Conf.~Computer Vision and Pattern
  Recognition}, pages 833--842, 2021.

\bibitem{zhang2017stackgan}
Han Zhang, Tao Xu, Hongsheng Li, Shaoting Zhang, Xiaogang Wang, Xiaolei Huang,
  and Dimitris~N Metaxas.
\newblock {StackGAN}: Text to photo-realistic image synthesis with stacked
  generative adversarial networks.
\newblock In {\em Proceedings of the IEEE international conference on computer
  vision}, pages 5907--5915, 2017.

\bibitem{zhang2023adding}
Lvmin Zhang and Maneesh Agrawala.
\newblock Adding conditional control to text-to-image diffusion models.
\newblock {\em arXiv preprint arXiv:2302.05543}, 2023.

\bibitem{zhu2019dm}
Minfeng Zhu, Pingbo Pan, Wei Chen, and Yi Yang.
\newblock {DM-GAN}: Dynamic memory generative adversarial networks for
  text-to-image synthesis.
\newblock In {\em Proc.~IEEE Int'l Conf.~Computer Vision and Pattern
  Recognition}, pages 5802--5810, 2019.

\end{thebibliography}
}

\end{document}